%% file: neurips_2026.tex
\documentclass{article}

\PassOptionsToPackage{numbers, square}{natbib}
\usepackage[preprint]{neurips_2026}

\usepackage[utf8]{inputenc} % allow utf-8 input
\usepackage[T1]{fontenc}    % use 8-bit T1 fonts
\usepackage{hyperref}       % hyperlinks
\usepackage{url}            % simple URL typesetting
\usepackage{booktabs}       % professional-quality tables
\usepackage{makecell}
\usepackage{amsfonts}       % blackboard math symbols
\usepackage{nicefrac}       % compact symbols for 1/2, etc.
\usepackage{microtype}      % microtypography
\usepackage[table]{xcolor}  
\usepackage{listings}
\usepackage{subcaption}
\usepackage{multirow}
\usepackage{amsmath,amssymb,amsthm}
\usepackage{mathtools}
\usepackage{tabularx}
\usepackage{hyperref}
\usepackage{graphicx}
\usepackage{algpseudocode}
\usepackage{wrapfig}
\usepackage{algorithm} 
\usepackage{xcolor}
\usepackage{listings}
\usepackage[T1]{fontenc}
\usepackage{xcolor}
\usepackage{tikz}
\usepackage[most]{tcolorbox}
\tcbuselibrary{skins,breakable}   
\definecolor{backcolour}{rgb}{0.95,0.95,0.92}
\definecolor{codegray}{rgb}{0.5,0.5,0.5}
\definecolor{codepurple}{rgb}{0.58,0,0.82}
\definecolor{violet}{rgb}{0.58,0,0.82}

\lstdefinestyle{mystyle}{
    backgroundcolor=\color{backcolour},
    commentstyle=\color{magenta},
    keywordstyle=\color{blue},
    numberstyle=\tiny\color{codegray},
    stringstyle=\color{codepurple},
    basicstyle=\fontfamily{\ttdefault}\footnotesize,
    breakatwhitespace=false,
    breaklines=true,
    keepspaces=true,
    frame=single,
    numbersep=5pt,
    showspaces=false,
    showstringspaces=false,
    showtabs=false,
    tabsize=2,
    classoffset=1,
    keywordstyle=\color{violet},
    classoffset=0,
}
\newtheorem{theorem}{Theorem}
\newtheorem{proposition}[theorem]{Proposition}

\newcommand{\ourmethod}{SCORE}
\title{Self-Evolving Deep Research via Joint Generation and Evaluation}

\author{%
  Han Zhu$^1$\thanks{Equal Contribution; $^\dag$Corresponding author.}, Chengkun Cai$^{2*}$, Yuanfeng Song$^{3\dag}$, Xing Chen$^3$, Sirui Han$^{1\dag}$, Yike Guo$^1$\\
  $^{1}$The Hong Kong University of Science and Technology\\
$^{2}$ByteDance, China $^{3}$ University College London
}

\begin{document}

\maketitle

\begin{abstract}\label{abstract}
Large Language Models (LLMs) have become increasingly adopted in daily applications, with deep research standing out as a particularly important capability. Unlike traditional question-answering (QA) tasks, deep research report generation lacks definitive ground-truth, making reward design inherently unverifiable and limiting effective reinforcement learning. Existing approaches mitigate this challenge with LLM-as-a-judge and query-dependent evaluation rubrics, but they still rely on static evaluators that cannot adapt their standards as the solver improves, leading to insufficient and eventually saturated optimization pressure. We address this limitation with a \textbf{s}elf-evolving \textbf{co}-evolutionary training framework for deep \textbf{re}search evaluation and generation (\textbf{\ourmethod}), which tightly couples an evaluator and a solver in a shared-parameter learning process. Rather than treating generation and evaluation as isolated modules, we leverage their intrinsic connection to enable joint improvement within a single shared-parameter model. To restrict this process, we introduce a meta-harness, which dynamically controls the evaluation environment based on solver performance, encouraging valid evaluation dimensions and sufficiently deep evaluator search. Extensive experiments on deep research benchmarks demonstrate consistent improvement in report generation quality, showing that co-evolving evaluation and generation is a promising direction for training open-ended research agents.
\end{abstract}

\input{files/introduction}

\input{files/related_work}

\input{files/methodology}

\input{files/experiment}

\section{Conclusion}
Generating comprehensive reports for deep research is a highly complex task, and the inherent difficulty of evaluating these open-ended reports severely hinders the training process. We observe a strong positive correlation between a model's evaluation and generation capabilities. Based on this insight, we propose {\ourmethod}, a self-evolving training framework that jointly optimizes an evaluator and a solver. Because the entire training procedure fundamentally relies on unverifiable rewards, we introduce an external meta-harness to impose necessary environmental constraints on the evaluator. This module simultaneously guides the evaluator to actively explore a broader range of relevant evaluation dimensions throughout the training phase. Finally, experimental results demonstrate that our approach achieves substantial performance improvements while requiring only a remarkably small amount of training data.

% \begin{ack}
% Use unnumbered first level headings for the acknowledgments. All acknowledgments

% Do {\bf not} include this section in the anonymized submission, only in the final paper. You can use the \texttt{ack} environment provided in the style file to automatically hide this section in the anonymized submission.
% \end{ack}

\bibliographystyle{plainnat}
\bibliography{custom}

%%%%%%%%%%%%%%%%%%%%%%%%%%%%%%%%%%%%%%%%%%%%%%%%%%%%%%%%%%%%
\newpage
\appendix
\input{files/appendix}

%%%%%%%%%%%%%%%%%%%%%%%%%%%%%%%%%%%%%%%%%%%%%%%%%%%%%%%%%%%%

% \newpage
% \input{checklist.tex}

\end{document}

%% file: files/introduction.tex
\section{Introduction \label{introduction}}
% 模型能力强，被广泛用于各种复杂任务，现有工作在 QA 任务上用 RL 取得了很好的效果
\begin{wrapfigure}{R}{0.5\textwidth}
    \centering
    \vspace{-15pt}
\includegraphics[width=0.5\textwidth]{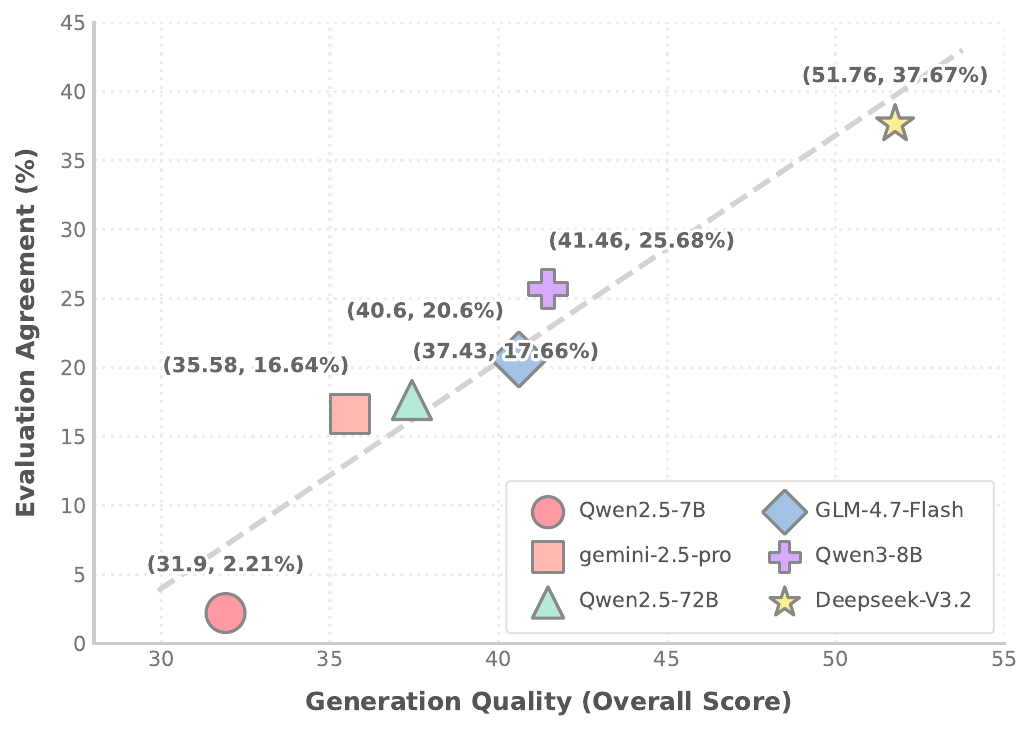}
    %\caption{LLMs demonstrate a positive correlation between generation and evaluation abilities.}
    \caption{Empirical evidence of the positive correlation between LLMs' generation and evaluation capabilities.}
    \label{gen_eval}
    \vspace{-10pt}
\end{wrapfigure}

Owing to their extraordinary capabilities, Large Language Models (LLMs) have become deeply integrated into various complex domains, spanning autonomous driving \citep{yang2025trajectoryllm}, software development \citep{dong2025survey}, multimedia generation \citep{seedance2026seedance,Yan_2026_WACV,zhang2026psgs} and so on \citep{yao2025countllm,li2026multiple}. As the demand for comprehensive information retrieval and synthesis grows, enhancing the deep research capabilities of LLMs has attracted increasing attention from researchers. Several search agents, such as GPT-Researcher \citep{gptresearcher} and AgentCPM-Explore \citep{chen2026agentcpmexplorerealizinglonghorizondeep}, have recently demonstrated remarkable proficiency in these multi-hop research tasks. Furthermore, previous works like Search-o1 \citep{li-etal-2025-search} and SSP \citep{lu2025search} have shown that applying Reinforcement Learning (RL) to these models can substantially enhance their reasoning capabilities when integrated into agent, yielding significant performance improvements on complex Question-Answering (QA) tasks.

% Report generation的2个challenge
Differ from traditional QA tasks, training models for open-ended report generation presents two challenges. First, comprehensive reports lack a definitive ground truth, making objective reward signals inherently difficult to define \citep{lei2026writingrladvancinglongformwriting,gao2026drarenaautomatedevaluationframework}. Since there is no absolute standard for open-ended queries, relying on human annotations for generated reports is prohibitively expensive, making objective and scalable reward signals difficult to construct. Relying on human annotations to cover these diverse paths is prohibitively expensive and difficult to scale. Second, the evaluation criteria for a high-quality report are multi-dimensional and query-related \citep{kim2024prometheus,du2026deepresearch}. Single fixed evaluation metric inevitably ignore crucial dimensions including factual correctness and insight diversity for different questions. This dimensional omission results in sparse training signals and fails to provide the fine-grained credit assignment required for complex generation tasks.

To overcome these challenges in report generation, recent research proposes several novel paradigms to provide more effective optimization signals. WebThinker \citep{li2026webthinker} and Step-DeepResearch \citep{hu2025stepdeepresearchtechnicalreport} have applied the LLM-as-a-judge \citep{zheng2023judging} to evaluate report quality during the training process. AdaRubric \citep{ding2026adarubrictaskadaptiverubricsllm} proposes a framework that dynamically generates task-specific evaluation criteria based on the input instructions. Mix-GRM \citep{zhang2026lengthscalingsynergizingbreadth} utilizes Reinforcement Learning with Verifiable Rewards (RLVR) to automatically adapt between breadth-first (B-CoT) and depth-first (D-CoT) evaluation strategies according to distinct task demands. However, a fixed external evaluator lacks the ability to dynamically adjust its evaluation standards, and therefore fails to provide sustained optimization pressure as the solver improves. 

% Our method
%As illustrated in Figure~\ref{gen_eval}, LLMs' generation quality is positively correlated with its evaluation capability. Based on this observation, we hypothesize that generation and evaluation abilities can be improved within a unified framework rather than being optimized in a decoupled manner. Therefore, we propose a self-play framework in which the policy model alternates between generating reports and evaluating them. By replacing a fixed external judge with a dynamic evaluator that evolves alongside the solver, our method maintains informative training pressure throughout learning. Furthermore, a Meta-Harness \citep{lee2026metaharnessendtoendoptimizationmodel} is applied to stabilize this process by enforcing environmental constraints. Empirically, this co-evolutionary paradigm brings substantial and consistent improvements across diverse research generation tasks.

As empirically observed in Figure~\ref{gen_eval}, LLMs' generation quality is positively correlated with its evaluation proficiency. 
Building upon this insight, we hypothesize that generation and evaluation abilities can be mutually reinforcing and should be optimized within a unified, rather than decoupled, architecture. To this end, we propose a self-play framework in which the policy model iteratively generates and evaluate reports. By replacing a fixed external judge with a dynamic evaluator that evolves alongside the solver, our method maintains informative training pressure throughout learning. Furthermore, a Meta-Harness \citep{lee2026metaharnessendtoendoptimizationmodel} is applied to stabilize this process by enforcing environmental constraints. Empirically, this co-evolutionary paradigm brings substantial and consistent improvements across diverse research generation tasks.

In a nutshell, our contributions are summarized as below:
\begin{itemize}
    \item We propose {\ourmethod}, an evaluator-solver self-play framework for training deep research agents with unverifiable rewards. This framework provide dynamic, multidimensional rewards to optimize the solver's ability in open-ended report generation, constrained by a meta-harness.
    \item We provide a theoretical analysis of {\ourmethod} under shared parameters. Our analysis clarifies the role of consistency in evaluator-side adaptation and characterizes the local optimization behavior of shared-parameter alternating updates.
    \item We show that our method improves agent performance across multiple evaluation dimensions on deep research benchmarks, and provide a systematic analysis of how parameters affect training procedure under unverifiable rewards.
\end{itemize}

%% file: files/related_work.tex
\section{Related Work}
\subsection{Deep Research}
LLMs have exhibited strong deep research capabilities in recent agent studies \citep{zhou2026lrasadvancedlegalreasoning,bai2026glance}. Initial frameworks, including Search-o1 \citep{li-etal-2025-search}, Search-R1 \citep{jin2025searchr}, and R1-Searcher \citep{song2025r1searcherincentivizingsearchcapability}, successfully introduced explicit reasoning process into agentic search frameworks, yielding substantial empirical improvement in multi-hop question-answering (QA) tasks. Following these advancements, researchers integrated diverse optimization algorithms into agentic training pipelines. Specifically, WebThinker \citep{li2026webthinker} utilizes iterative Direct Preference Optimization (DPO) for deep research agents, whereas DeepResearcher \citep{zheng-etal-2025-deepresearcher} adopts Group Relative Policy Optimization (GRPO). Diverging from these algorithmic enhancements, SimpleDeepSearcher \citep{sun-etal-2025-simpledeepsearcher} argues that the quality of the training data supersedes the specific training paradigm. 
Beyond QA tasks, report generation tasks are inherently more complex, as they involve open-ended, multi-faceted synthesis where establishing reliable ground truth for automated evaluation remains a critical bottleneck. To address this complexity, multi-agent frameworks like STORM and Co-STORM leverage preliminary knowledge and human-in-the-loop interactions during report generation \citep{shao-etal-2024-assisting,jiang-etal-2024-unknown}. Later approaches such as Step-DeepResearch and AgentCPM-Report \citep{hu2025stepdeepresearchtechnicalreport,li2026agentcpmreportinterleavingdraftingdeepening} directly integrate model training optimized for their specific agent frameworks.

\subsection{Intrinsic Rewards and Self-Play}
Traditional Self-play is widely applied to agentic studies as an adversarial reinforcement learning approach \citep{zhang2024survey}. AbsoluteZero, R-Zero, SeRL and SPIRAL, demonstrate the effectiveness of self-evolution in tackling complex reasoning tasks under data scarcity situation \citep{zhao2025absolute, huang2025r, liu2025spiral, fang2026serl}. Recent frameworks, including Search Self-play, Dr. Zero, and SPaR, demonstrate the substantial advantages of applying this technique to search agents operating within multi-hop environments \citep{lu2025search, yue2026dr, cheng2025sparselfplaytreesearchrefinement}. Beyond adversarial learning paradigms like proposer-solver self-play, alternative approaches propose training solvers on filtered trajectories generated by the models themselves, as demonstrated by frameworks such as WebThinker, EvolveSearch, and EXSEARCH \citep{li2026webthinker, zhang-etal-2025-evolvesearch, shi2026iterative}. Other notable efforts include AceSearcher, which employs cooperative interaction between decomposer and solver for self-improvement \citep{xu2026acesearcher}. Moreover, optimization strategies utilizing internal rewards provided by models themselves have been adopted by researchers in frameworks such as INTUITOR \citep{zhao2026learning} and JEPO \citep{tang2026beyond}. These previous works demonstrate the efficacy of internal signals in enhancing autonomous capability improvements. However, the unique characteristics fo deep research task present a significant evaluation challenge. The difficulty of reliably scoring open-domain, comprehensive reports makes it highly challenging to design effective reward signals that resolve the credit assignment issue.

%% file: files/methodology.tex
\section{Methodology}\label{methodology}
\begin{figure}[t]
    \centering
    \includegraphics[width=1\linewidth]{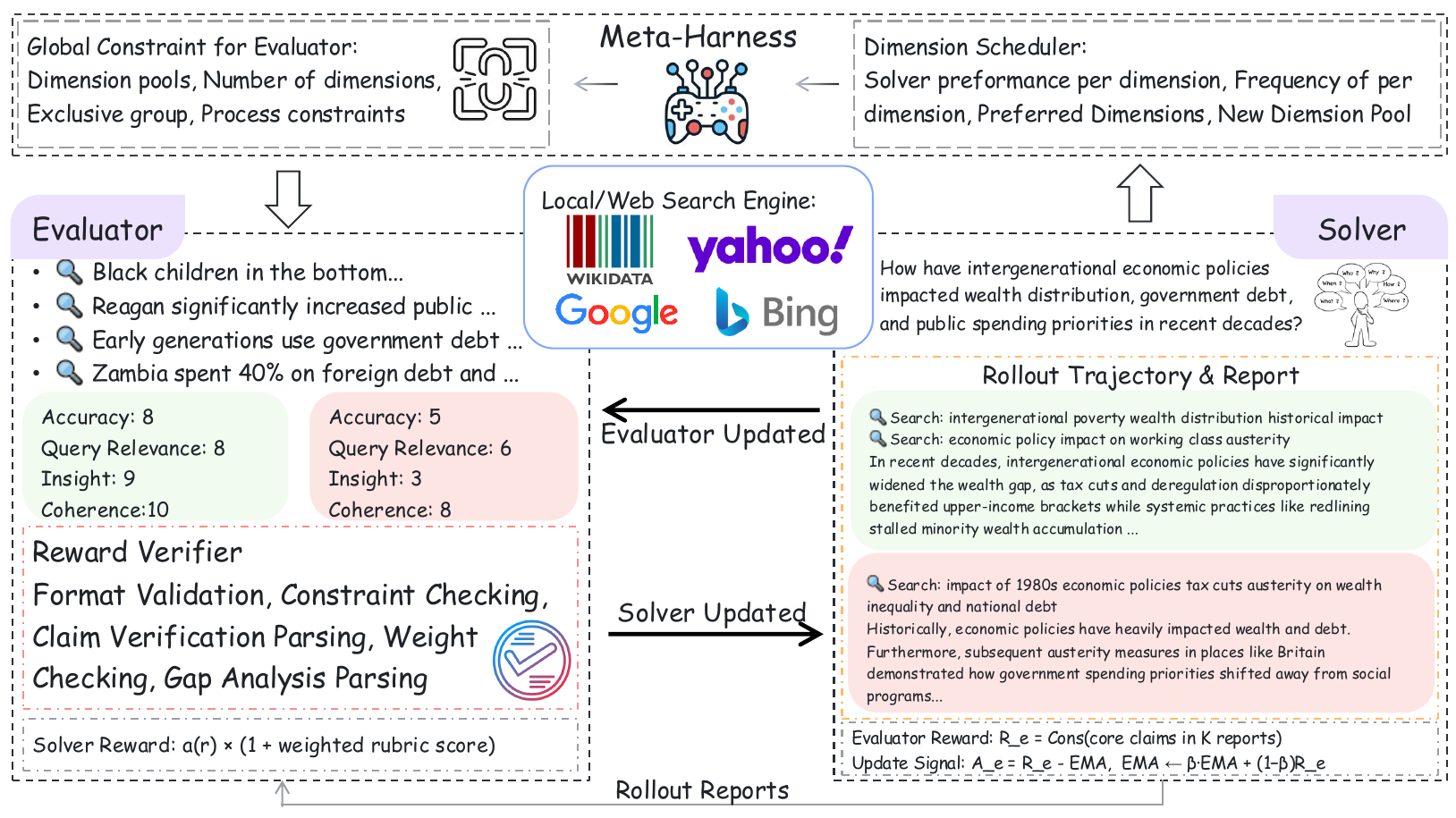}
    \caption{{\ourmethod} is controlled by Meta-Harness. The evaluator first selects evaluation dimensions from a provided dimension pool and retrieves evidence accordingly. The solver is updated using rewards derived from evaluator, while the evaluator is optimized based on the consistency of rollout reports generated by the solver. A reward verifier filters out invalid rewards before they are used for solver optimization.}
    \label{pipeline}
    \vspace{-15pt}
\end{figure}
\subsection{Problem Setting}
\label{sec:problem}

We study \emph{deep research report generation}, where the goal is to learn a policy
$
\pi(r \mid q, \mathcal{E}_q)
$
over candidate reports $r$ for a given query $q$ and environment $\mathcal{E}_q$. We assume that report quality is governed by a latent functional
$
Q(r; q, \mathcal{E}_q),
$
and the ideal objective is to assign more probability mass to reports with higher latent quality. Here, $q \in \mathcal{Q}$ denotes a user query, $r \in \mathcal{R}$ denotes a candidate long-form report, and $\mathcal{E}_q$ denotes the query-specific evidence and evaluation environment.

Unlike QA-style tasks, $Q$ is inherently \emph{multi-dimensional} and \emph{query-dependent}: a report may succeed or fail along different axes, such as factual support, evidence coverage, citation grounding, contradiction handling, uncertainty calibration, and discourse coherence. As a result, long-form report generation rarely admits a single reliable verifier or a fixed scalar reward that is both sufficiently informative and broadly applicable.

We therefore formulate deep research as a query-conditioned report optimization problem with an evaluation bottleneck. In particular, the central challenge in training deep research models is how to evaluate long-form reports in a query-specific and stable manner under limited supervision.

\subsection{Overview of {\ourmethod}}
\label{sec:overview}

To address this challenge, we propose {\ourmethod}, a retrieval-grounded and externally guided evaluator-solver framework for deep research report generation as shown in Figure~\ref{pipeline}. {\ourmethod} contains four components:
\begin{enumerate}
    \item an external evidence environment that can be queried through search and retrieval tools,
    \item a fixed external \emph{meta-harness} that shapes the query-specific evaluation environment,
    \item an \emph{evaluator} that constructs query-conditioned rubrics and produces structured report assessments,
    \item a \emph{solver} that interacts with the evidence environment, gathers supporting evidence, and generates reports under the resulting rubric.
\end{enumerate}

A key design choice of {\ourmethod} is that the solver and evaluator are implemented as two functionally distinct roles over a \emph{shared underlying actor}. This design is motivated by the substantial overlap between report generation and report evaluation, both of which require query understanding, evidence modeling, and discourse-level reasoning. We empirically examine this choice in our ablations Section~\ref{ablation_study}.

Training proceeds through sequential alternating updates on the shared actor: the solver is optimized using evaluator-derived report rewards, while the evaluator is optimized using rubric-level auxiliary signals. Both updates are regularized by KL divergence against a reference policy to limit excessive policy drift.

\subsection{External Evidence and Query-Specific Environment}
\label{sec:environment}

For each query $q$, {\ourmethod} interacts with an external evidence environment through search and retrieval tools. We denote by
\[
D_q^{(0)} = \mathcal{R}(q)
\]
the initial evidence obtained from this environment, where $\mathcal{R}$ abstracts the retrieval interaction available to the system. In practice, this evidence environment may include web search, document retrieval, or other tool-mediated evidence sources.

Based on the initial evidence $D_q^{(0)}$, {\ourmethod} constructs a query-specific evaluation environment through a fixed external \emph{meta-harness}:
\[
\mathcal{E}_q = \mathrm{Harness}(q, D_q^{(0)}, \mathcal{S}),
\]
where $\mathcal{S}$ denotes recent training statistics or historical signals available to the harness. The resulting environment $\mathcal{E}_q$ specifies the admissible evaluation space for the current query, including candidate evaluation dimensions, structural constraints on valid rubrics, process requirements for evidence verification or comparison, and auxiliary task-specific guidance.

The meta-harness is external and non-trainable. It is implemented as a fixed external controller and is not optimized by gradient descent together with the actor. Its role is to shape the admissible evaluation space in which evaluator-solver interaction takes place.

\subsection{Evaluator and Solver}
\label{sec:roles}

Given query $q$ and environment $\mathcal{E}_q$, the evaluator samples a query-conditioned rubric
\[
(H_q, w_q) \sim \pi_e(\cdot \mid q, \mathcal{E}_q),
\]
where $H_q=\{h_1,\dots,h_k\}$ denotes the selected evaluation dimensions and $w_q$ their corresponding weights.

Conditioned on the query, the current evidence state, and the evaluator-selected rubric, the solver interacts with the external evidence environment to gather supporting information and generate a group of candidate reports
\[
r_1,\dots,r_K \sim \pi_s(\cdot \mid q, D_q^{s}, H_q, w_q),
\]
where $D_q^{s}$ denotes the evidence collected by the solver and $K$ is the number of solver rollouts sampled for the same query during training.

For assessment, the evaluator may perform constrained supplementary retrieval or verification based on the query and the generated reports, producing its own evidence context $D_q^{e}$. It then produces structured judgments over candidate reports, including dimension-wise scoring, evidence support checking, contradiction analysis, and citation verification. These structured judgments are converted into solver-side learning signals.

\subsection{Shared Actor and Training Signals}
\label{sec:signals}

{\ourmethod} implements the solver and evaluator as two roles over a shared actor with parameters $\theta$. This design is motivated by the observation that report generation and report evaluation rely on overlapping latent capabilities, including query understanding, evidence modeling, and discourse-level reasoning. Under this view, generation and evaluation are not treated as fully separate functions, but as two role-specific mappings induced from a shared underlying representation. A shared actor therefore provides a common representational substrate for both roles, allowing evaluator-side adaptation to influence the same internal representations used by the solver. We empirically examine this design choice in Section~\ref{ablation_study}.

{\ourmethod} uses two distinct reward channels:
\[
\underbrace{R_s(r; q, D_q^{e}, H_q, w_q)}_{\text{solver-side report reward}}
\qquad \text{and} \qquad
\underbrace{R_e(\Gamma_q)=C(\Gamma_q)}_{\text{evaluator-side auxiliary reward}}.
\]
The solver-side reward supervises report generation, while the evaluator-side reward adapts the evaluator's rubric-selection behavior.

\paragraph{Solver-side report reward.}
For each generated report $r$, the evaluator produces a scalar report reward
\[
R_s(r; q, D_q^{e}, H_q, w_q),
\]
where $D_q^{e}$ denotes the evidence context used by the evaluator for rubric-conditioned assessment, $H_q=\{h_1,\dots,h_m\}$ denotes the selected evaluation dimensions, and $w_q=\{w_1,\dots,w_m\}$ denotes their corresponding weights. This reward serves as the primary supervision signal for optimizing the solver.

In our implementation, the abstract report reward is instantiated as
\[
R_s(r)
=
a(r)\cdot
\left(
1+\sum_{j=1}^{m} w_j\,s_j(r;q,D_q^{e})
\right),
\]
where $a(r)$ denotes a report-level validity or accuracy score, and $s_j(r;q,D_q^{e})$ denotes the evaluator score of report $r$ on dimension $h_j$ under evaluator-side evidence context $D_q^{e}$. Intuitively, the solver is encouraged to generate reports that are both valid and strong along the evaluator-selected rubric dimensions.

\paragraph{Evaluator-side auxiliary reward.}
To adapt the evaluator, {\ourmethod} introduces a rubric-level auxiliary reward based on inter-report consistency. For a fixed query and evaluator-selected rubric
\[
\Gamma_q := (H_q,w_q),
\]
the solver generates $K$ reports under the same query and evaluation setting. We define the raw evaluator-side auxiliary reward as
\[
R_e(\Gamma_q)=C(\Gamma_q),
\]
where $C(\Gamma_q)$ is the empirical consistency score computed by aggregating pairwise agreement across the generated reports. In the current implementation, $C(\Gamma_q)$ is computed as the batch mean of pairwise agreement scores over extracted report conclusions.

Evaluator updates use an EMA-centered version of this signal,
\[
A_e(\Gamma_q)=R_e(\Gamma_q)-b_{t-1},
\]
where $b_{t-1}$ is the previous exponential moving average baseline of past auxiliary rewards. After computing the centered signal, the baseline is updated as
\[
b_t
=
\beta b_{t-1}
+
(1-\beta)R_e(\Gamma_q),
\]
where $\beta$ is the EMA decay coefficient. Thus, $R_e$ favors rubrics under which solver behavior is more reproducible across repeated rollouts, while $A_e$ provides the actual lower-variance signal used for REINFORCE-style evaluator updates.

Overall, {\ourmethod} forms a two-channel training loop: evaluator-derived report rewards improve the solver, while consistency-based rubric-level rewards adapt the evaluator. The solver-side reward targets report quality under the selected rubric, whereas the evaluator-side reward favors rubrics that induce more stable and reproducible solver behavior. In this sense, consistency is used as a stability-oriented auxiliary signal for rubric adaptation rather than as a direct correctness criterion.

\subsection{Sequential Alternating KL-Regularized Optimization}
\label{sec:optimization}

Given the two reward channels defined in Section~\ref{sec:signals}, training in {\ourmethod} proceeds through sequential alternating updates on the shared actor. Each step constructs a query-specific evaluation environment, generates rubric-conditioned reports, evaluates them under evaluator-side verification context, and then applies solver-side and evaluator-side updates to the shared parameters $\theta$.

The solver-side update uses evaluator-derived report rewards $R_s$, which are converted into group-normalized advantages and optimized with GRPO. The evaluator-side update uses the EMA-centered auxiliary reward $A_e$, derived from the raw consistency reward $R_e$, to update the rubric-selection policy with a REINFORCE-style objective. Although these updates are computed from different trajectories and rewards, both act on the same actor and are therefore coupled through the shared parameterization.

Both updates are regularized by KL divergence against a fixed reference policy $\pi_{\mathrm{ref}}$. In our implementation, this KL term is applied at the actor-update level rather than as an explicit reward penalty. Thus, KL regularization is not treated as a third reward channel, but as an optimization regularizer that limits excessive policy drift under shared-actor learning.

At the analysis level, we denote the unregularized role-specific objectives by $\mathcal{L}_s(\theta)$ and $\mathcal{L}_e(\theta)$, respectively. The corresponding KL-regularized objectives can be written as
\[
\mathcal{L}_s^{\mathrm{reg}}(\theta)
=
\mathcal{L}_s(\theta)
+
\lambda\,\mathrm{KL}\!\left(\pi_\theta \,\|\, \pi_{\mathrm{ref}}\right),
\]
\[
\mathcal{L}_e^{\mathrm{reg}}(\theta)
=
\mathcal{L}_e(\theta)
+
\lambda\,\mathrm{KL}\!\left(\pi_\theta \,\|\, \pi_{\mathrm{ref}}\right),
\]
where $\lambda > 0$ is the KL regularization coefficient.

KL regularization serves two purposes in {\ourmethod}. First, it stabilizes alternating optimization by limiting abrupt policy drift under shared-role learning. Second, it helps maintain compatibility between solver-side and evaluator-side updates on the shared actor. In this way, {\ourmethod} combines evaluator-derived report rewards, consistency-based evaluator adaptation, external grounding, shared-role learning, and KL-regularized alternating optimization into a unified training framework for long-form report generation.

\section{Theoretical Analysis} \label{theory}
\label{sec:theory}

We analyze two properties of {\ourmethod}: the role of consistency as an evaluator-side auxiliary signal, and the first-order optimization behavior of shared-parameter KL-regularized alternating updates.

\paragraph{Assumptions.}
We assume that: (A1) the analysis-level objectives $\mathcal{L}_s$ and $\mathcal{L}_e$ are $L$-smooth; (A2) the corresponding stochastic gradients are unbiased or approximately unbiased with bounded variance; and (A3) the stochastic gradients are almost surely bounded.

\paragraph{Consistency as rubric-induced reproducibility.}
Let $\Gamma_q := (H_q, w_q)$ denote the evaluator-selected rubric for query $q$. Suppose the solver generates $K$ reports under the same query, evidence context, and rubric, and let
$
C(\Gamma_q)
$
denote the resulting empirical consistency score obtained by aggregating pairwise agreement scores across these reports. In the current implementation, this is the mean pairwise consistency across generated reports. We define the raw evaluator-side auxiliary signal as
\[
R_e(\Gamma_q)=C(\Gamma_q),
\]
and use the centered signal
\[
A_e(\Gamma_q)=R_e(\Gamma_q)-b_t
\]
for evaluator updates, where $b_t$ is an exponential moving average baseline.

\begin{proposition}[Consistency estimates rubric-induced reproducibility]
\label{prop:consistency_short}
Let $\mu_C(\Gamma_q)$ denote the population mean of the consistency metric under rubric $\Gamma_q$. Then
\[
\mathbb{E}[R_e(\Gamma_q)]
=
\mathbb{E}[\mu_C(\Gamma_q)].
\]
Thus, the raw evaluator-side auxiliary signal estimates the expected reproducibility induced by rubric $\Gamma_q$ under repeated solver rollouts.
\end{proposition}

Higher reproducibility suggests that the rubric induces a more stable and repeatable evaluation context for the solver, which makes it a useful auxiliary signal for evaluator-side adaptation. Centering by the EMA baseline reduces gradient variance without changing the underlying preference encoded by the raw signal. In {\ourmethod}, consistency is therefore used as a rubric-level auxiliary signal for evaluator-side adaptation.

\paragraph{First-order characterization of alternating updates.}
Let the shared-actor updates be
\[
\theta' = \theta_t - \eta_s \hat g_s(\theta_t),
\qquad
\theta_{t+1} = \theta' - \eta_e \hat g_e(\theta'),
\]
where $\hat g_s$ and $\hat g_e$ are the stochastic gradients of the solver-side and evaluator-side objectives, respectively. Define the analysis surrogate
\[
\widetilde{\mathcal{L}}(\theta)
=
\mathcal{L}_s(\theta)
+
\alpha \mathcal{L}_e(\theta)
+
\lambda\,\mathrm{KL}\!\left(\pi_\theta \,\|\, \pi_{\mathrm{ref}}\right),
\qquad
\alpha=\eta_e/\eta_s,
\]
where $\pi_{\mathrm{ref}}$ is the fixed reference policy and $\lambda>0$ is the KL regularization coefficient.

\begin{proposition}[First-order characterization of alternating updates]
\label{prop:alternating_short}
Under Assumptions (A1)--(A3), the expected one-step {\ourmethod} update satisfies
\[
\mathbb{E}[\theta_{t+1}-\theta_t \mid \theta_t]
=
-\eta_s \nabla \widetilde{\mathcal{L}}(\theta_t)+\rho_t,
\qquad
\|\rho_t\| = O(\eta_s^2).
\]
\end{proposition}

This result shows that the shared-parameter solver-evaluator updates follow, to first order, a KL-regularized surrogate descent direction, with only a higher-order residual induced by sequential alternating updates. Detailed derivations are deferred to Appendix~\ref{app:theory}.

%% file: files/experiment.tex
\section{Experiments}
\subsection{Experimental Setup} \label{experimental_setup}
\paragraph{Training Details} 
\input{files/main_results}
We train all models using the VeRL framework \citep{Sheng_2025}. The evaluator and solver are updated asynchronously that the solver is updated at every training step using GRPO \citep{DBLP:journals/corr/abs-2402-03300}, while the evaluator is updated once every 1 steps using REINFORCE \citep{williams1992simple}. We set the rollout number to 1 for the evaluator and 8 for the solver. The maximum number of tool calls is 10, the batch size is 64, the maximum response length is 4,096 tokens, and the maximum input length is 8,192 tokens. For training, we use a local search engine backed by local WikiData, while web search is used during evaluation. GPT-5.2 \citep{gpt52} is applied as Meta-Harness and is invoked every 5 training steps to modify the environment according to the solver’s current performance. For the coefficient of EMA baseline, we set $\beta$=0.7. We construct the training set by collecting and preprocessing user queries from Reddit. 

\paragraph{Baselines and Models}
We implement {\ourmethod} on top of two capable base LLMs, Qwen2.5-7B-Instruct \citep{qwen2025qwen25technicalreport} and Llama-3.1-8B-Instruct \citep{grattafiori2024llama3herdmodels}. For comparison, we train these models using standard GRPO and DPO to serve as baselines. The post-trained models are subsequently evaluated within two representative search agent systems to assess their downstream capabilities. Specifically, we employ open-deep-research \citep{opendeepresearch} as a baseline for the ReAct paradigm and gpt-researcher \citep{gptresearcher} as a baseline for the plan-and-execute paradigm. To maintain experimental consistency, a single, unified model is deployed across all functional modules within these respective agents.

\paragraph{Benchmarks and Evaluation Metrics}
To rigorously evaluate our approach, we employ two benchmarks designed for deep research tasks which are DeepResearchBench \citep{du2026deepresearch} and DeepResearchEval \citep{app16052546}. DeepResearchBench assesses agent performance across multi-dimensions including comprehensiveness, insight, instruction following, readability, and citation validity. In contrast, DeepResearchEval utilizes a dynamic evaluation metric, which aligns closely with our methodological perspective. Both frameworks ultimately adopt GPT-5.2 \citep{zheng2023judging} for the final evaluation. 

\subsection{Experimental Results}
\paragraph{Main Results}
As illustrated in Table~\ref{main_result}, {\ourmethod} demonstrates a substantial and robust improvement in analytical depth and valid citation across all agents. Basic reinforcement learning (RL) methods such as GRPO and DPO suffer from a catastrophic degradation in at lease one dimension, occasionally collapsing to exactly zero in Plan-and-Execute agents when integrating LLama-3.1-8B-Instruct. This phenomenon is driven by Lazy Likelihood Displacement (LLD), a condition where the likelihood of correct reasoning trajectories decays during tool-integrated RL, ultimately causing training collapse \citep{deng2026on}. Furthermore, agent architectural magnify this instability, as decoupled long-horizon frameworks like Plan-and-Execute exhibit premature collapse and severe reward hacking compared to step-by-step ReAct policies \citep{deng2026grouprelativepolicyoptimization}. In contrast, {\ourmethod} maintains stable performance by utilizing a co-evolving evaluator that provides dense, consistent constraints through dynamic evaluation rubrics. Notably, we observe a consistent decline in readability across nearly all training methods. We hypothesis this a fundamental trade-off because models are optimized to synthesize comprehensive analysis and rigorously cite evidence, the generated report inherently incorporates denser academic vocabulary and more complex logical structures, thereby sacrificing conversational simplicity for analytical rigor which might be similar to the trade-off between diversity and factuality \citep{li-etal-2024-dawn}.

\begin{wrapfigure}{R}{0.45\textwidth}
    \centering
    \vspace{-15pt}
\includegraphics[width=0.45\textwidth]{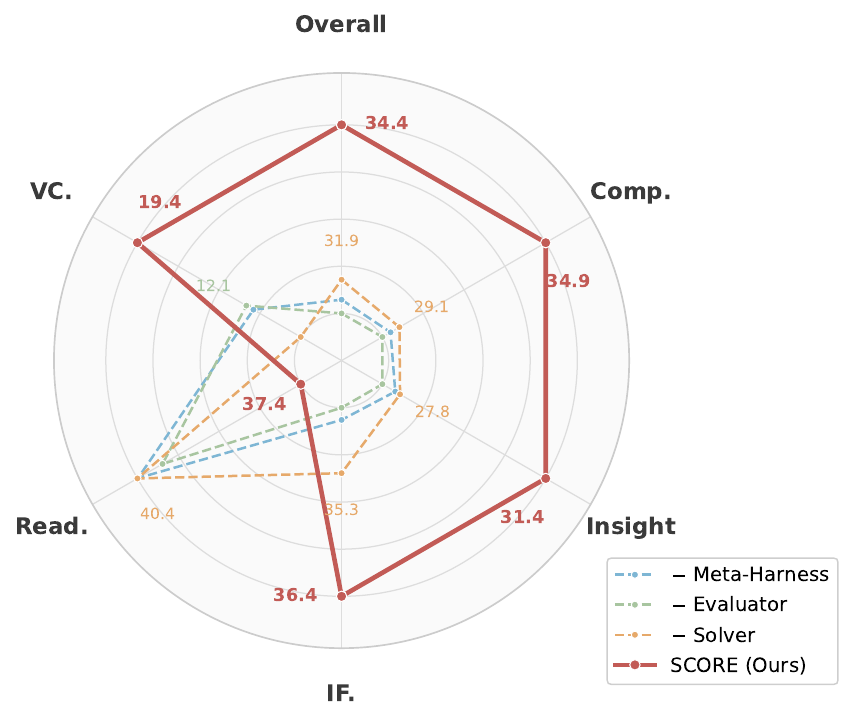}
    \caption{Ablation study for individual modules on DeepResearchBench. \textbf{Comp.}: comprehensivenss, \textbf{IF}: instruction following, \textbf{Read.}: Readability, \textbf{VC}: valid ratio of citation.}
    \label{module_ablation}
    \vspace{-15pt}
\end{wrapfigure}
\paragraph{Module Ablation Study \label{ablation_study}} 
Refer to Figure~\ref{module_ablation}, we observe that removing any single module severely degrades the LLMs' comprehensiveness, demonstrating that generating well-rounded research reports requires the synergistic operation of the entire framework. Freezing the solver leads to a massive decline in citation validity. The internal consistency reward for evaluator brings reward hacking, where the evaluator learns to construct superficial rubrics that elicit highly consistent text from the static solver while actively avoiding high-variance behaviors like rigorous external citation. Furthermore, our complete method exhibits a slightly lower readability score compared to the ablated baselines, which we attribute to the inherent trade-off between natural language coherence and academic formatting. Base models tend to produce fluent but unstructured text, whereas our co-evolutionary framework forces the model to generate deeper reports with complex citation syntax, thus sacrificing superficial readability for factual rigor.

\paragraph{Experiments on Rollout Number} 
\begin{figure}[htbp]
    \centering
    % 第一张子图
    \begin{subfigure}{0.33\textwidth}
        \centering
        \includegraphics[width=\textwidth]{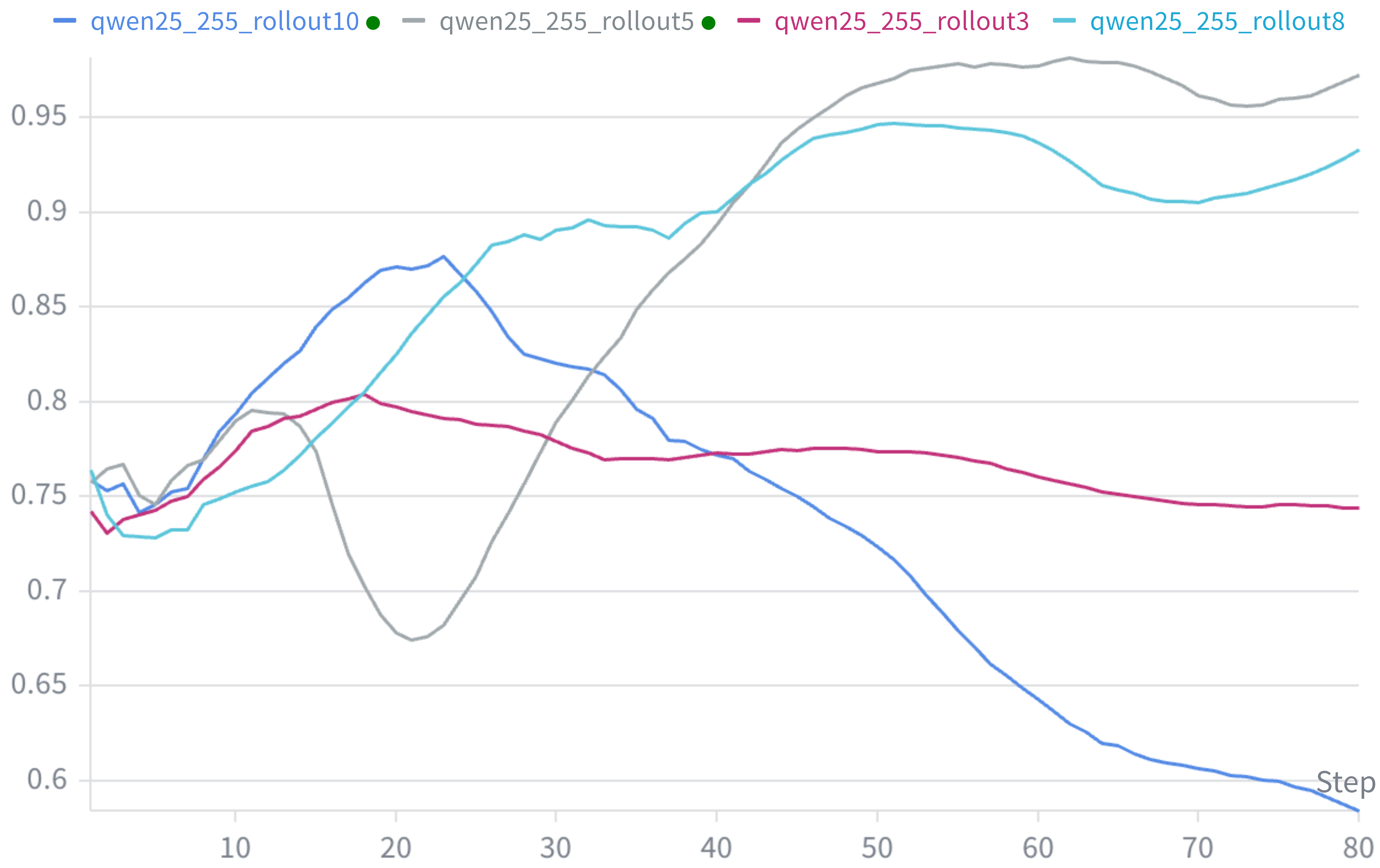} % 替换为你的图片文件名
        \caption{Solver Entropy}
    \end{subfigure}\hfill
    % 第二张子图
    \begin{subfigure}{0.33\textwidth}
        \centering
        \includegraphics[width=\textwidth]{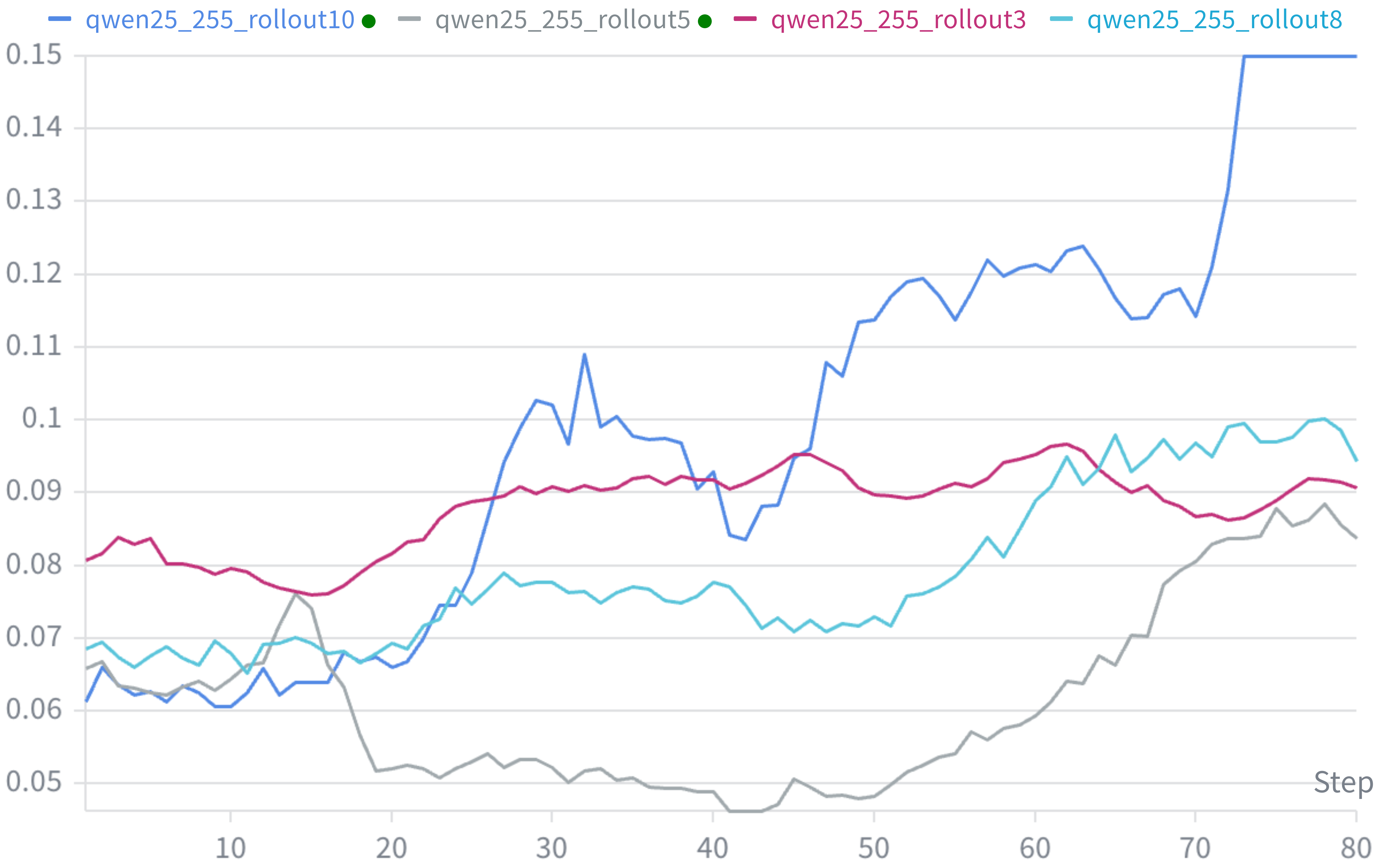}
        \caption{Evaluator Reward}
    \end{subfigure}\hfill
    % 第三张子图
    \begin{subfigure}{0.33\textwidth}
        \centering
        \includegraphics[width=\textwidth]{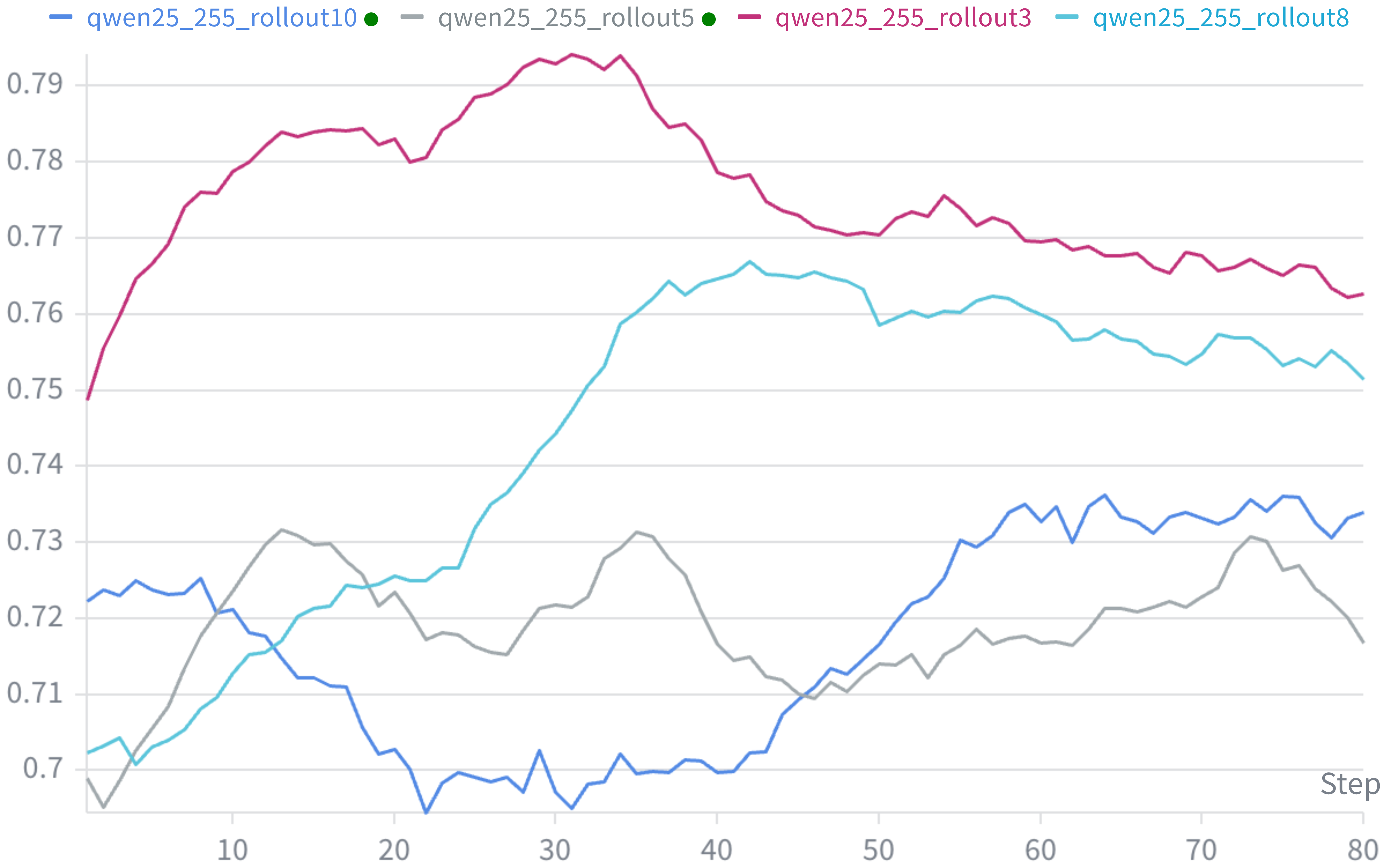}
        \caption{Solver Reward}
    \end{subfigure}\hfill
    \caption{Training Details}
    \label{rollout_ablation}
    \vspace{-10pt}
\end{figure}

We investigate the impact of the rollout number $K$ in the training progress. Figure~\ref{rollout_ablation} demonstrates that extreme $K$ affect training procedure  that small $K=3$ fails to yield meaningful reward improvements, while large $K=10$ triggers a severe model collapse reflected by sudden drop of entropy \citep{shiintrinsic}. Small $K$ fails to provide the solver with statistically robust reward due to insufficient ranking samples while for the evaluator, fewer rollouts yield unreliable pairwise consistency estimates, resulting in noisy REINFORCE gradients that prevent stable dimension policy updates. Conversely, an overly large $K$ yields highly precise advantage estimates for both the solver and the evaluator, accelerating their updates. Under the self-evolving setting, this mutual acceleration compounds policy drift and may trigger rapid model collapse.

%\subsection{Case Study}
\paragraph{Case Study}

As illustrated in Figure~\ref{case1} in Appendix~\ref{additional_experiments}, different types of research queries inherently require distinct evaluation criteria. For instance, legal queries demand absolute precision and strict citation of statutory texts, whereas open-ended questions require the aggregation of diverse perspectives to maintain fairness and objectivity. A generic, static rubric fails to capture these critical nuances, often assigning acceptable scores to reports that are superficially fluent but fundamentally flawed. In contrast, our adaptive evaluation framework dynamically introduces specialized evaluation dimensions tailored to each query. This mechanism ensures that the generated reports are accurately penalized or rewarded according to the specific rigor demanded by the task.

%% file: files/main_results.tex
\begin{table}[t] 
    \centering
    \caption{Performance of open-deep-research and gpt-researcher integrated with LLaMA-3.1 and Qwen2.5 on DeepResearchBench. \textbf{Comp.}: comprehensivenss, \textbf{IF}: instruction following, \textbf{Read.}: Readability, \textbf{VC}: valid ratio of citation.}
    \label{main_result}
    
    % 排版紧凑化设置
    \renewcommand{\arraystretch}{1.1}
    \setlength{\tabcolsep}{3pt} % 缩小列间距，节省横向空间
    
    \resizebox{\textwidth}{!}{%
        \begin{tabular}{l ccccc c}
        \toprule
        % \rowcolor{blue!10}
        & \multicolumn{5}{c}{\textbf{Writing Quality}} & \\
        \cmidrule(lr){2-6}
        % \rowcolor{blue!10}
        \multirow{-2}{*}{\textbf{Models and Agents}} & \textbf{Overall} & \textbf{Comp.} & \textbf{Insight} & \textbf{IF.} & \textbf{Read.} & \multirow{-2}{*}{\makecell{\textbf{VC.}}} \\
        \midrule
        
        % ================= 分组 1: Open-deep-research =================
        \multicolumn{7}{l}{\textit{\textcolor{gray}{Agent: Open-deep-research (ReAct Paradigm)}}} \\
        
        % LLaMA 组
        \rowcolor{gray!5}
        \quad Llama-3.1-8B-Inst.     & 27.00 & 24.05 & 22.46 & 30.28 & 36.29 & 22.58 \\
        \quad \quad + GRPO & 
        23.41 {\fontsize{7}{7}\selectfont \textcolor{red!70!black}{(-3.59)}} & 
        19.89 {\fontsize{7}{7}\selectfont \textcolor{red!70!black}{(-4.16)}} & 
        18.73 {\fontsize{7}{7}\selectfont \textcolor{red!70!black}{(-3.73)}} & 
        26.77 {\fontsize{7}{7}\selectfont \textcolor{red!70!black}{(-3.51)}} & 
        33.90 {\fontsize{7}{7}\selectfont \textcolor{red!70!black}{(-2.39)}} & 
        28.57 {\fontsize{7}{7}\selectfont \textcolor{green!50!black}{(+5.99)}} \\
        \rowcolor{gray!5}
        \quad \quad + DPO  & 
        26.13 {\fontsize{7}{7}\selectfont \textcolor{red!70!black}{(-0.87)}} & 
        23.11 {\fontsize{7}{7}\selectfont \textcolor{red!70!black}{(-0.94)}} & 
        21.28 {\fontsize{7}{7}\selectfont \textcolor{red!70!black}{(-1.18)}} & 
        29.84 {\fontsize{7}{7}\selectfont \textcolor{red!70!black}{(-0.44)}} & 
        35.77 {\fontsize{7}{7}\selectfont \textcolor{red!70!black}{(-0.52)}} & 
        28.00  {\fontsize{7}{7}\selectfont \textcolor{green!50!black} {(+5.42)}} \\
        \rowcolor{blue!5}
        \quad \quad + {\ourmethod} (Ours) & 
        30.07 {\fontsize{7}{7}\selectfont \textcolor{green!50!black}{(+3.07)}} & 
        30.31 {\fontsize{7}{7}\selectfont \textcolor{green!50!black}{(+6.26)}} & 
        26.71 {\fontsize{7}{7}\selectfont \textcolor{green!50!black}{(+4.25)}} & 
        32.25 {\fontsize{7}{7}\selectfont \textcolor{green!50!black}{(+1.97)}} & 
        33.66 {\fontsize{7}{7}\selectfont \textcolor{red!70!black}{(-2.63)}} & 
        30.55 {\fontsize{7}{7}\selectfont \textcolor{green!50!black}{(+7.97)}} \\
        
        \addlinespace[2pt]
        
        % Qwen 组
        \rowcolor{gray!5}
        \quad Qwen2.5-7B-Inst.       & 31.92 & 28.93 & 27.81 & 35.37 & 40.29 & 11.76 \\
        \quad \quad + GRPO & 
        32.67 {\fontsize{7}{7}\selectfont \textcolor{green!50!black}{(+0.75)}} & 
        29.94 {\fontsize{7}{7}\selectfont \textcolor{green!50!black}{(+1.01)}} & 
        29.08 {\fontsize{7}{7}\selectfont \textcolor{green!50!black}{(+1.27)}} & 
        35.30 {\fontsize{7}{7}\selectfont \textcolor{red!70!black}{(-0.07)}} & 
        40.87 {\fontsize{7}{7}\selectfont \textcolor{green!50!black}{(+0.58)}} & 
        16.13 {\fontsize{7}{7}\selectfont \textcolor{green!50!black}{(+4.37)}} \\
        \rowcolor{gray!5}
        \quad \quad + DPO  & 
        33.91 {\fontsize{7}{7}\selectfont \textcolor{green!50!black}{(+1.99)}} & 
        34.26 {\fontsize{7}{7}\selectfont \textcolor{green!50!black}{(+5.33)}} & 
        30.89 {\fontsize{7}{7}\selectfont \textcolor{green!50!black}{(+3.08)}} & 
        35.90 {\fontsize{7}{7}\selectfont \textcolor{green!50!black}{(+0.53)}} & 
        36.85 {\fontsize{7}{7}\selectfont \textcolor{red!70!black}{(-3.44)}} & 
        13.76 {\fontsize{7}{7}\selectfont \textcolor{green!50!black}{(+2.00)}} \\
        \rowcolor{blue!5}
        \quad \quad + {\ourmethod} (Ours) & 
        34.43 {\fontsize{7}{7}\selectfont \textcolor{green!50!black}{(+2.51)}} & 
        34.86 {\fontsize{7}{7}\selectfont \textcolor{green!50!black}{(+5.93)}} & 
        31.38 {\fontsize{7}{7}\selectfont \textcolor{green!50!black}{(+3.57)}} & 
        36.37 {\fontsize{7}{7}\selectfont \textcolor{green!50!black}{(+1.00)}} & 
        37.41 {\fontsize{7}{7}\selectfont \textcolor{red!70!black}{(-2.88)}} & 
        19.35 {\fontsize{7}{7}\selectfont \textcolor{green!50!black}{(+7.59)}} \\
        
        \midrule
        
        % ================= 分组 2: gpt-researcher =================
        \multicolumn{7}{l}{\textit{\textcolor{gray}{Agent: gpt-researcher (Plan-and-Execute Paradigm)}}} \\
        
        % LLaMA 组
        \rowcolor{gray!5}
        \quad Llama-3.1-8B-Inst.     & 28.92 & 25.62 & 24.80 & 31.85 & 38.36 & 38.23  \\
        \quad \quad + GRPO & 
        16.97 {\fontsize{7}{7}\selectfont \textcolor{red!70!black}{(-11.95)}} & 
        16.79 {\fontsize{7}{7}\selectfont \textcolor{red!70!black}{(-38.23)}} & 
        13.12 {\fontsize{7}{7}\selectfont \textcolor{red!70!black}{(-11.68)}} & 
        18.09 {\fontsize{7}{7}\selectfont \textcolor{red!70!black}{(-13.76)}} & 
        23.78 {\fontsize{7}{7}\selectfont \textcolor{red!70!black}{(-14.58)}} & 
        0.00  {\fontsize{7}{7}\selectfont \textcolor{red!70!black}{(-8.16)}} \\
        \rowcolor{gray!5}
        \quad \quad + DPO  & 
        33.58 {\fontsize{7}{7}\selectfont \textcolor{green!50!black}{(+4.66)}} & 
        33.83 {\fontsize{7}{7}\selectfont \textcolor{green!50!black}{(+8.21)}} & 
        30.64 {\fontsize{7}{7}\selectfont \textcolor{green!50!black}{(+5.84)}} & 
        35.15 {\fontsize{7}{7}\selectfont \textcolor{green!50!black}{(+3.30)}} &
        36.87 {\fontsize{7}{7}\selectfont \textcolor{red!70!black}{(-1.49)}} & 
        34.48 {\fontsize{7}{7}\selectfont \textcolor{red!70!black}{(-3.75)}}\\
        \rowcolor{blue!5}
        \quad \quad + {\ourmethod} (Ours) & 
         33.89 {\fontsize{7}{7}\selectfont \textcolor{green!50!black}{(+4.97)}} & 
         34.14 {\fontsize{7}{7}\selectfont \textcolor{green!50!black}{(+8.52)}} & 
         30.90 {\fontsize{7}{7}\selectfont \textcolor{green!50!black}{(+6.10)}} & 
         35.64 {\fontsize{7}{7}\selectfont \textcolor{green!50!black}{(+3.79)}} & 
         37.29 {\fontsize{7}{7}\selectfont \textcolor{red!70!black}{(-1.07)}} & 
        52.38 {\fontsize{7}{7}\selectfont \textcolor{green!50!black}{(+14.15)}}\\
        
        \addlinespace[2pt]
        
        % Qwen 组
        \rowcolor{gray!5}
        \quad Qwen2.5-7B-Inst.       & 32.00 & 29.20 & 28.70 & 34.67 & 39.61 & 36.11 \\
        \quad \quad + GRPO & 
        30.03 {\fontsize{7}{7}\selectfont \textcolor{red!70!black}{(-1.97)}} & 
        27.52 {\fontsize{7}{7}\selectfont \textcolor{red!70!black}{(-1.68)}} & 
        27.39 {\fontsize{7}{7}\selectfont \textcolor{red!70!black}{(-1.31)}} & 
        31.07 {\fontsize{7}{7}\selectfont \textcolor{red!70!black}{(-3.60)}} & 
        38.37 {\fontsize{7}{7}\selectfont \textcolor{red!70!black}{(-1.24)}} & 
        51.61 {\fontsize{7}{7}\selectfont \textcolor{green!50!black}{(+15.50)}} \\
        \rowcolor{gray!5}
        \quad \quad + DPO  & 
        35.85 {\fontsize{7}{7}\selectfont \textcolor{green!50!black}{(+3.85)}} & 
        36.22 {\fontsize{7}{7}\selectfont \textcolor{green!50!black}{(+7.02)}} & 
        33.64 {\fontsize{7}{7}\selectfont \textcolor{green!50!black}{(+4.94)}} & 
        37.15 {\fontsize{7}{7}\selectfont \textcolor{green!50!black}{(+2.48)}} & 
        38.07 {\fontsize{7}{7}\selectfont \textcolor{red!70!black}{(-1.54)}} & 
        30.80 {\fontsize{7}{7}\selectfont \textcolor{red!70!black}{(-5.31)}} \\
        \rowcolor{blue!5}
        \quad \quad + {\ourmethod} (Ours) & 
        35.91 {\fontsize{7}{7}\selectfont \textcolor{green!50!black}{(+3.91)}} & 
        36.15 {\fontsize{7}{7}\selectfont \textcolor{green!50!black}{(+6.95)}} & 
        33.12 {\fontsize{7}{7}\selectfont \textcolor{green!50!black}{(+4.42)}} & 
        37.60 {\fontsize{7}{7}\selectfont \textcolor{green!50!black}{(+2.93)}} & 
        38.93 {\fontsize{7}{7}\selectfont \textcolor{red!70!black}{(-0.68)}} & 
        66.15 {\fontsize{7}{7}\selectfont \textcolor{green!50!black}{(+30.04)}}\\
        
        \bottomrule
        \end{tabular}%
    }
   % \vspace{-20pt}
\end{table}

%% file: files/appendix.tex
\section{Limitations \label{limitation}}
Despite the effectiveness of our self-evolving framework, we acknowledge several limitations in our current study. First, the unique nature of deep research report generation demands exceptional instruction-following capabilities from the base language models. Models with weaker foundational abilities may struggle with dynamic rubrics generation and tool call. Second, our training paradigm relies heavily on entirely unverifiable rewards. This lack of ground-truth makes the model susceptible to policy collapse during the optimization process. Consequently, careful hyperparameter tuning and strict early stopping mechanisms are necessary to ensure training stability. Even though our framework successfully provides differentiated reward signals for specific criteria, aggressively optimizing for one dimension can sometimes lead to performance degradation in another. Balancing these competing objectives simultaneously remains a challenging problem for future exploration.

\section{Future Work}
According to our experiments, we observe that inherent trade-offs still exist during the training process for deep research report generation. We hypothesize that this issue is primarily caused by sparse reward signals resulting from exceptionally long agent trajectories. To address this challenge, future work will explore phased training paradigms. By decomposing the deep research generation process into distinct functional stages and assigning intermediate rewards, we aim to resolve the long-standing credit assignment problem that plagues end-to-end policy optimization.

\section{LLM Usage Declaration} \label{llm_declaration}
LLMs are employed to improve the grammar and language quality of the paper. LLMs are also applied to check the correctness and clarity of the proofs. In the experiments, open-source LLMs were used for post-train models, while closed-source LLMs were used as judges for assessment. In addition, a coding agent was used to assist in developing parts of the training framework. 

\section{Licenses} \label{license}
We use the following models in the experiments:
\begin{itemize}
    \item Qwen2.5-7B-Instruct, Qwen2.5-72B-Instruct \citep{qwen2025qwen25technicalreport} and Qwen3-8B \citep{yang2025qwen3technicalreport} with License \textit{Apache 2.0}.
    \item Llama-3.1-8B-Instruct \citep{grattafiori2024llama3herdmodels} with License \textit{Llama 3.1 Community}.
    \item GLM-4.7-Flash \citep{5team2025glm45agenticreasoningcoding} and DeepSeek-V3.2 \citep{deepseekai2025deepseekv32pushingfrontieropen} with License \textit{mit}.
    \item Gemini-2.5-pro \citep{comanici2025gemini25pushingfrontier} with License \textit{Apache 2.0} and GPT-5.2 \citep{gpt52} with proprietary license.
    
\end{itemize}

\section{Algorithm}

Algorithm~\ref{alg:score_main} summarizes the main shared-actor evaluator--solver training loop, while Algorithm~\ref{alg:score_eval} details the evaluator-side consistency reward construction and rubric adaptation step. In the current implementation, \textsc{PairwiseConsistency} is instantiated as the mean pairwise TF-IDF cosine similarity over extracted rollout conclusions.

\begin{algorithm}[H]
\caption{{\ourmethod}: Evaluator--Solver Self-Play with Consistency-Driven Dimension Learning}
\label{alg:score_main}
\begin{algorithmic}[1]
\Require Shared actor parameters $\theta$; evaluator role policy $\pi_e(\cdot;\theta)$; solver role policy $\pi_s(\cdot;\theta)$;
dimension pool $\mathcal{D}$; solver rollout size $K$; batch size $B$;
evaluator update period $U$; harness refresh period $M$; minimum verification/search depth $\tau_s$; KL coefficient $\lambda_{\mathrm{KL}}$
\Ensure Trained shared actor parameters $\theta$

\State Initialize DimensionScheduler $\mathcal{S}$ over $\mathcal{D}$
\State Initialize Meta-Harness state $\mathcal{M}_0 \gets \mathcal{M}_{\mathrm{default}}$
\State Initialize evaluator-side EMA baseline $b_0 \gets 0$

\For{each training step $t = 1,2,\dots,T_{\max}$}

    \Statex \hspace{\algorithmicindent}\textcolor{gray}{$\triangleright$ \emph{Phase 1: Refresh scheduler and harness constraints}}
    \State $(\mathcal{R}_t,\mathcal{A}_t) \gets \mathcal{S}.\textsc{GetConstraints}()$
    \Comment{retrieve the current required / avoided dimensions and exploration hints}
    \State Obtain current dimension and process constraints from $\mathcal{M}_{t-1}$
    \Comment{the meta-harness defines the admissible rubric space for this step}

    \Statex \hspace{\algorithmicindent}\textcolor{gray}{$\triangleright$ \emph{Phase 2: Evaluator samples a rubric}}
    \State Sample query batch $\{q_i\}_{i=1}^{B}$
    \State Build evaluator rubric-generation prompt $x_t^{\mathrm{rubric}}$ from $(\mathcal{R}_t,\mathcal{A}_t,\mathcal{M}_{t-1})$
    \State Sample rubric
    \[
    (H_t,w_t),\ \log \pi_e(H_t,w_t \mid x_t^{\mathrm{rubric}};\theta)
    \sim \pi_e(\cdot \mid x_t^{\mathrm{rubric}};\theta)
    \]
    \Comment{the evaluator chooses what dimensions to emphasize and how to weight them}

    \Statex \hspace{\algorithmicindent}\textcolor{gray}{$\triangleright$ \emph{Phase 3: Solver performs rubric-conditioned rollouts}}
    \For{each query $q_i$, each rollout $k = 1,\dots,K$}
        \State Generate report trajectory and final report
        \[
        y_i^k \sim \pi_s(\cdot \mid q_i, H_t, w_t;\theta)
        \]
        \State Record solver-side evidence / tool observations $D_{q_i,k}^{s}$
        \Comment{the solver searches, interacts with tools, and produces a long-form report}
    \EndFor

    \Statex \hspace{\algorithmicindent}\textcolor{gray}{$\triangleright$ \emph{Phase 4: Evaluator assesses reports}}
    \State Optionally build evaluator-side evidence contexts $\{D_{q_i}^{e}\}_{i=1}^{B}$ through constrained retrieval / verification
    \Comment{the evaluator may gather its own evidence for rubric-conditioned assessment}
    \For{each query $q_i$, each rollout $k = 1,\dots,K$}
        \State Compute evaluator-side report reward
        \[
        s_i^k \gets \textsc{EvalScore}(q_i, y_i^k, H_t, w_t, D_{q_i}^{e}, \mathcal{M}_{t-1}; \theta)
        \]
        \State \textbf{if} evaluator verification/search depth for $(i,k)$ is below $\tau_s$ \textbf{then} set $s_i^k \gets 0$
        \Comment{discard under-verified rewards to avoid noisy solver supervision}
    \EndFor

    \Statex \hspace{\algorithmicindent}\textcolor{gray}{$\triangleright$ \emph{Phase 5: Solver-side GRPO update}}
    \For{each query $q_i$}
        \State $\mu_i \gets \operatorname{mean}_{k} s_i^k,\quad \sigma_i \gets \operatorname{std}_{k} s_i^k$
        \State $\hat A_i^k \gets (s_i^k-\mu_i)/(\sigma_i+\varepsilon)$
        \Comment{group-normalized rewards define solver-side advantages}
    \EndFor
    \State Update shared actor parameters $\theta$ using the solver-side GRPO objective with KL regularization against $\pi_{\mathrm{ref}}$

    \Statex \hspace{\algorithmicindent}\textcolor{gray}{$\triangleright$ \emph{Phase 6: Evaluator-side consistency update}}
    \State Compute the evaluator-side consistency signal and update the evaluator role as detailed in Algorithm~\ref{alg:score_eval}
    \Comment{the evaluator is rewarded when its rubric induces more reproducible solver behavior}

    \Statex \hspace{\algorithmicindent}\textcolor{gray}{$\triangleright$ \emph{Phase 7: Refresh scheduler and meta-harness}}
    \State $\mathcal{S}.\textsc{UpdateStats}(H_t, \{C_i\}_{i=1}^{B})$
    \Comment{record which dimensions produced more reproducible rollouts}
    \If{$t \bmod M = 0$}
        \State Collect harness statistics $\mathcal{T}_t$
        \State $\mathcal{M}_t \gets \textsc{Harness.Generate}(\mathcal{T}_t, \mathcal{M}_{t-1}, \mathcal{M}_{\mathrm{default}})$
        \Comment{periodically refresh the admissible evaluation space}
    \Else
        \State $\mathcal{M}_t \gets \mathcal{M}_{t-1}$
    \EndIf
\EndFor
\end{algorithmic}
\end{algorithm}

\begin{algorithm}[H]
\caption{Evaluator-Side Consistency Reward and Rubric Update}
\label{alg:score_eval}
\begin{algorithmic}[1]
\Require Rollout reports $\{y_i^k\}_{i=1,\dots,B}^{k=1,\dots,K}$; sampled rubric $(H_t,w_t)$; rubric prompt $x_t^{\mathrm{rubric}}$;
previous EMA baseline $b_{t-1}$; baseline decay $\beta$; evaluator scale $\alpha$; evaluator update period $U$
\Ensure Updated EMA baseline $b_t$, consistency statistics $\{C_i\}_{i=1}^{B}$, and optionally updated shared actor parameters $\theta$

\Statex \hspace{\algorithmicindent}\textcolor{gray}{$\triangleright$ \emph{Phase 1: Build a batch-level consistency signal}}
\For{each query $q_i$}
    \State Extract rollout conclusions $\{c_i^k\}_{k=1}^{K}$ from $\{y_i^k\}_{k=1}^{K}$
    \Comment{measure agreement at the conclusion level rather than at the token level}
    \State Compute empirical consistency
    \[
    C_i \gets \textsc{PairwiseConsistency}(\{c_i^k\}_{k=1}^{K})
    \]
\EndFor
\State Compute evaluator-side raw signal
\[
R_{\mathrm{eval}} \gets \frac{1}{B}\sum_{i=1}^{B} C_i
\]
\Comment{the raw signal favors rubrics under which solver rollouts are more reproducible}

\Statex \hspace{\algorithmicindent}\textcolor{gray}{$\triangleright$ \emph{Phase 2: Center the signal with an EMA baseline}}
\State Compute centered evaluator-side signal
\[
\hat A_{\mathrm{eval}} \gets R_{\mathrm{eval}} - b_{t-1}
\]
\State Update EMA baseline
\[
b_t \gets \beta\, b_{t-1} + (1-\beta)\, R_{\mathrm{eval}}
\]
\Comment{the EMA baseline stabilizes REINFORCE updates without changing the raw preference}

\Statex \hspace{\algorithmicindent}\textcolor{gray}{$\triangleright$ \emph{Phase 3: Update the evaluator role}}
\If{$t \bmod U = 0$ \textbf{and} evaluator updates are enabled}
    \State Update shared actor parameters $\theta$ using the evaluator-side REINFORCE objective
    \[
    \theta \gets \theta + \eta\, \alpha\, \hat A_{\mathrm{eval}}\,
    \nabla_\theta \log \pi_e(H_t,w_t \mid x_t^{\mathrm{rubric}};\theta)
    \]
    \Comment{rubrics that induce more reproducible solver behavior become more likely}
\EndIf
\end{algorithmic}
\end{algorithm}

\section{Detailed Proofs}
% \appendix
\label{app:theory}

This appendix provides derivations for the results stated in Section~\ref{sec:theory}. Throughout, let $\theta \in \mathbb{R}^d$ denote the shared actor parameters, and let
\[
\widetilde{\mathcal{L}}(\theta)
=
\mathcal{L}_s(\theta)
+
\alpha \mathcal{L}_e(\theta)
+
\lambda\,\mathrm{KL}\!\left(\pi_\theta \,\|\, \pi_{\mathrm{ref}}\right)
\]
denote the KL-regularized surrogate potential.

\subsection{Derivation of Proposition~\ref{prop:consistency_short}}
\label{app:proof_consistency}

Recall that $\Gamma_q := (H_q,w_q)$ denotes the evaluator-selected rubric for query $q$. Suppose the solver generates $K$ reports under the same query, evidence context, and rubric. Let
\[
c_{ij}(\Gamma_q), \qquad 1 \le i < j \le K,
\]
denote the pairwise agreement score between reports $r_i$ and $r_j$ under rubric $\Gamma_q$, and let
\[
M=\binom{K}{2}
\]
be the number of report pairs. In the current implementation, the empirical consistency score is defined as
\[
C(\Gamma_q)
=
\frac{1}{M}\sum_{1 \le i < j \le K} c_{ij}(\Gamma_q).
\]
The raw evaluator-side auxiliary signal is
\[
R_e(\Gamma_q)=C(\Gamma_q).
\]

Let
\[
\mu_C(\Gamma_q):=\mathbb{E}[c_{ij}(\Gamma_q)]
\]
denote the population mean of the consistency metric under rubric $\Gamma_q$. Then
\[
\mathbb{E}[R_e(\Gamma_q)]
=
\mathbb{E}[C(\Gamma_q)]
=
\mathbb{E}\!\left[
\frac{1}{M}\sum_{1 \le i < j \le K} c_{ij}(\Gamma_q)
\right].
\]
By linearity of expectation,
\[
\mathbb{E}[R_e(\Gamma_q)]
=
\frac{1}{M}\sum_{1 \le i < j \le K}\mathbb{E}[c_{ij}(\Gamma_q)]
=
\frac{1}{M}\sum_{1 \le i < j \le K}\mu_C(\Gamma_q)
=
\mu_C(\Gamma_q).
\]
Taking expectation over the induced rubric distribution yields
\[
\mathbb{E}[R_e(\Gamma_q)]
=
\mathbb{E}[\mu_C(\Gamma_q)].
\]
This proves Proposition~\ref{prop:consistency_short}.

In implementation, evaluator updates use the centered signal
\[
A_e(\Gamma_q)=R_e(\Gamma_q)-b_t,
\]
where $b_t$ is an exponential moving average baseline. Since $b_t$ is independent of the current sampled rubric action, subtracting it reduces gradient variance without changing the underlying expected preference encoded by $R_e(\Gamma_q)$.

\subsection{Derivation of Proposition~\ref{prop:alternating_short}}
\label{app:proof_alternating}

The shared-actor updates are
\[
\theta' = \theta_t - \eta_s \hat g_s(\theta_t),
\qquad
\theta_{t+1} = \theta' - \eta_e \hat g_e(\theta').
\]
Hence,
\[
\theta_{t+1}-\theta_t
=
-\eta_s \hat g_s(\theta_t)-\eta_e \hat g_e(\theta').
\]
Taking conditional expectation with respect to $\theta_t$ gives
\[
\mathbb{E}[\theta_{t+1}-\theta_t \mid \theta_t]
=
-\eta_s \mathbb{E}[\hat g_s(\theta_t)\mid \theta_t]
-\eta_e \mathbb{E}[\hat g_e(\theta')\mid \theta_t].
\]
Under Assumption (A2),
\[
\mathbb{E}[\hat g_s(\theta_t)\mid \theta_t] \approx \nabla \mathcal{L}_s(\theta_t),
\qquad
\mathbb{E}[\hat g_e(\theta')\mid \theta_t] \approx \nabla \mathcal{L}_e(\theta').
\]
Thus,
\[
\mathbb{E}[\theta_{t+1}-\theta_t \mid \theta_t]
\approx
-\eta_s \nabla \mathcal{L}_s(\theta_t)
-\eta_e \nabla \mathcal{L}_e(\theta').
\]

By Assumption (A1), $\mathcal{L}_e$ is $L$-smooth, so
\[
\|\nabla \mathcal{L}_e(\theta')-\nabla \mathcal{L}_e(\theta_t)\|
\le
L\|\theta'-\theta_t\|.
\]
From the first update,
\[
\theta'-\theta_t = -\eta_s \hat g_s(\theta_t),
\]
and by Assumption (A3), $\|\hat g_s(\theta_t)\|\le G$. Therefore,
\[
\|\theta'-\theta_t\| \le \eta_s G,
\]
which implies
\[
\|\nabla \mathcal{L}_e(\theta')-\nabla \mathcal{L}_e(\theta_t)\|
\le
L\eta_s G.
\]
Hence we may write
\[
\nabla \mathcal{L}_e(\theta')
=
\nabla \mathcal{L}_e(\theta_t)+\delta_t,
\qquad
\|\delta_t\| = O(\eta_s).
\]

Substituting this into the expected update yields
\[
\mathbb{E}[\theta_{t+1}-\theta_t \mid \theta_t]
=
-\eta_s \nabla \mathcal{L}_s(\theta_t)
-\eta_e \nabla \mathcal{L}_e(\theta_t)
-\eta_e \delta_t.
\]
Using $\alpha=\eta_e/\eta_s$, we obtain
\[
\mathbb{E}[\theta_{t+1}-\theta_t \mid \theta_t]
=
-\eta_s \nabla \bigl(\mathcal{L}_s(\theta_t)+\alpha \mathcal{L}_e(\theta_t)\bigr)
+\rho_t^{(0)},
\]
where
\[
\rho_t^{(0)}=-\eta_e \delta_t,
\qquad
\|\rho_t^{(0)}\| = O(\eta_s^2).
\]

Including KL regularization against the fixed reference policy $\pi_{\mathrm{ref}}$, we define
\[
\widetilde{\mathcal{L}}(\theta)
=
\mathcal{L}_s(\theta)
+
\alpha \mathcal{L}_e(\theta)
+
\lambda\,\mathrm{KL}\!\left(\pi_\theta \,\|\, \pi_{\mathrm{ref}}\right).
\]
The same argument then gives
\[
\mathbb{E}[\theta_{t+1}-\theta_t \mid \theta_t]
=
-\eta_s \nabla \widetilde{\mathcal{L}}(\theta_t)+\rho_t,
\qquad
\|\rho_t\| = O(\eta_s^2),
\]
which proves Proposition~\ref{prop:alternating_short}.

\subsection{Summary}
\label{app:theory_summary}

The above derivations establish two facts used in the main text:
\begin{enumerate}
    \item the raw evaluator-side consistency signal estimates the expected reproducibility induced by a rubric under repeated solver rollouts;
    \item the shared-parameter solver-evaluator alternating updates follow, to first order, a KL-regularized surrogate descent direction, with only a higher-order residual due to sequential updating.
\end{enumerate}

\section{Experiments}
\subsection{Training Details} \label{additional_experimental_setup}
\begin{table}[h]
\centering
\footnotesize
\setlength{\tabcolsep}{4pt}
\renewcommand{\arraystretch}{1.08}
\caption{Main training hyperparameters in {\ourmethod}.}
\label{tab:training_details}
\begin{tabularx}{\columnwidth}{>{\raggedright\arraybackslash}p{0.36\columnwidth} >{\raggedright\arraybackslash}X}
\toprule
\textbf{Category} & \textbf{Configuration} \\
\midrule
Learning rate & $1\times10^{-6}$  \\
Warmup & 5 steps, warmup ratio 0.05 \\
Training Data Size & 255 for Qwen2.5-7B and 2,380 for Llama-3.1-8B \\
KL regularization & Coefficient 0.05, low-variance KL, applied in actor loss only \\
Batch size & Training batch size 64; PPO mini-batch size 32; PPO micro-batch size 4 per GPU \\
Rollouts & Solver rollout number 8; evaluator rollout number 1 \\
Context and generation limits & Maximum input length 8192; maximum response length 4096; evaluator maximum generation length 2048 \\
Interaction budget & Maximum assistant turns 10 \\
Evaluator update &  EMA baseline decay 0.7; update frequency every step; temperature 0.8 \\
Meta-harness & Refresh frequency every 5 training steps \\
Training retrieval & Local static retrieval backend use wikidata \\
Compute & Single-node training with 4 H800, 100 CPU cores, and 1 TB host memory \\
Systems optimization & FSDP, Ray orchestration, gradient checkpointing, parameter offloading, optimizer offloading \\
Training schedule & 50-120 steps\\
\bottomrule
\end{tabularx}
\end{table}

\subsection{Experimental Results} \label{additional_experiments}
\paragraph{Results of DeepResearchEval}
\input{files/deepresearcheval}
We also evaluate the search agents on DeepResearchEval \citep{app16052546}, where evaluation rubrics are customized for each specific query. The results shown in Table~\ref{deepresearcheval} strongly corroborate our previous observations that fundamental GRPO methods suffer severe performance degradation, whereas our framework consistently achieves the highest scores across all dimensions, including the highly challenging query-specific Meta metrics. Interestingly, while generic readability declines on DeepResearchBench benchmarks, our method achieves consistent gains in the ``Clarity'' metric on DeepResearchEval. This divergence supports our trade-off hypothesis, confirming that the generated texts exchange conversational simplicity for a highly structured and analytically clear presentation, which is heavily rewarded when evaluating complex, query-specific research explanations.

\begin{figure}[htbp]
    \centering
    % 第一张子图
    \begin{subfigure}{0.33\textwidth}
        \centering
        \includegraphics[width=\textwidth]{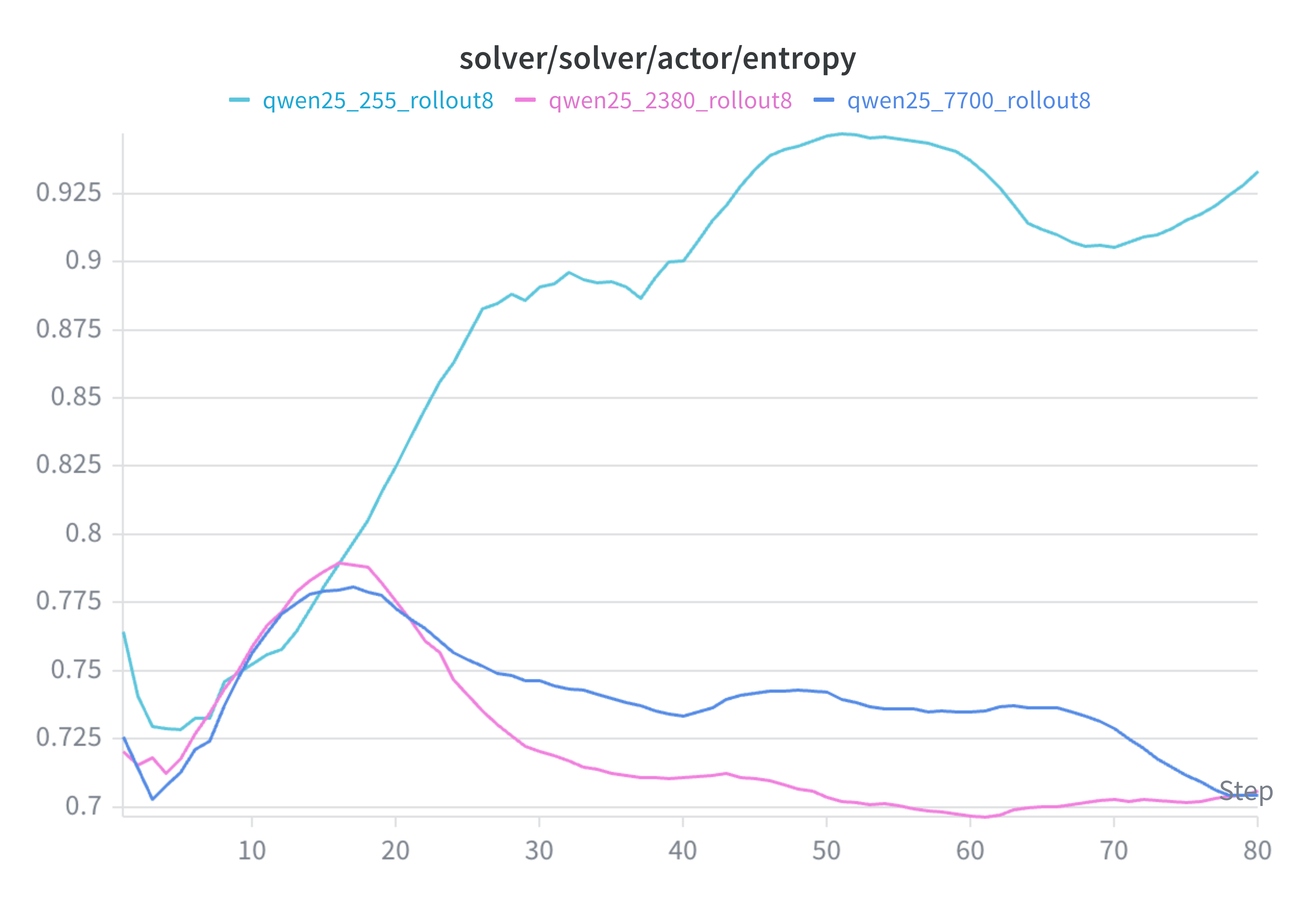} % 替换为你的图片文件名
        \caption{Solver Entropy}
    \end{subfigure}\hfill
    % 第二张子图
    \begin{subfigure}{0.33\textwidth}
        \centering
        \includegraphics[width=\textwidth]{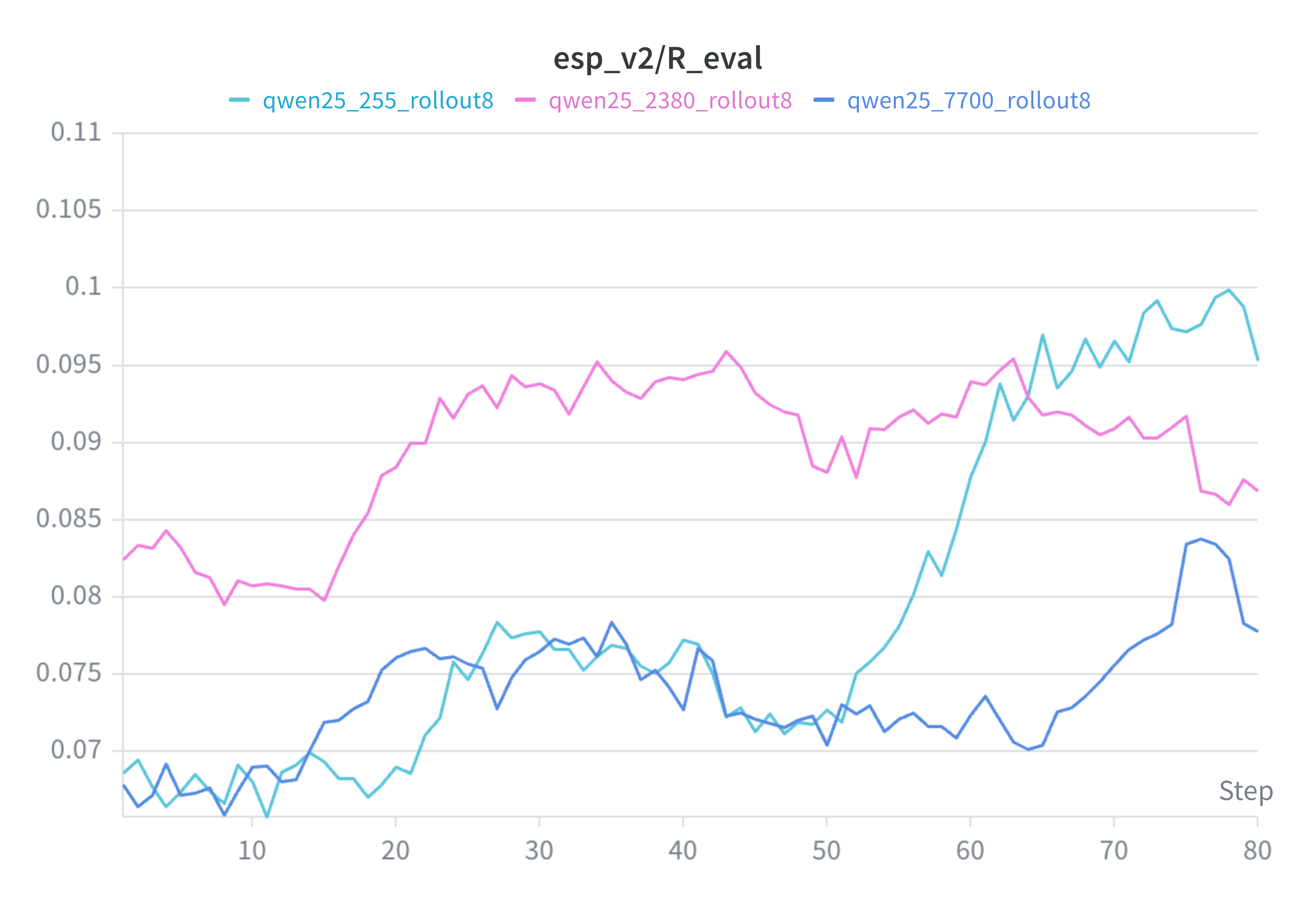}
        \caption{Evaluator Reward}
    \end{subfigure}\hfill
    % 第三张子图
    \begin{subfigure}{0.33\textwidth}
        \centering
        \includegraphics[width=\textwidth]{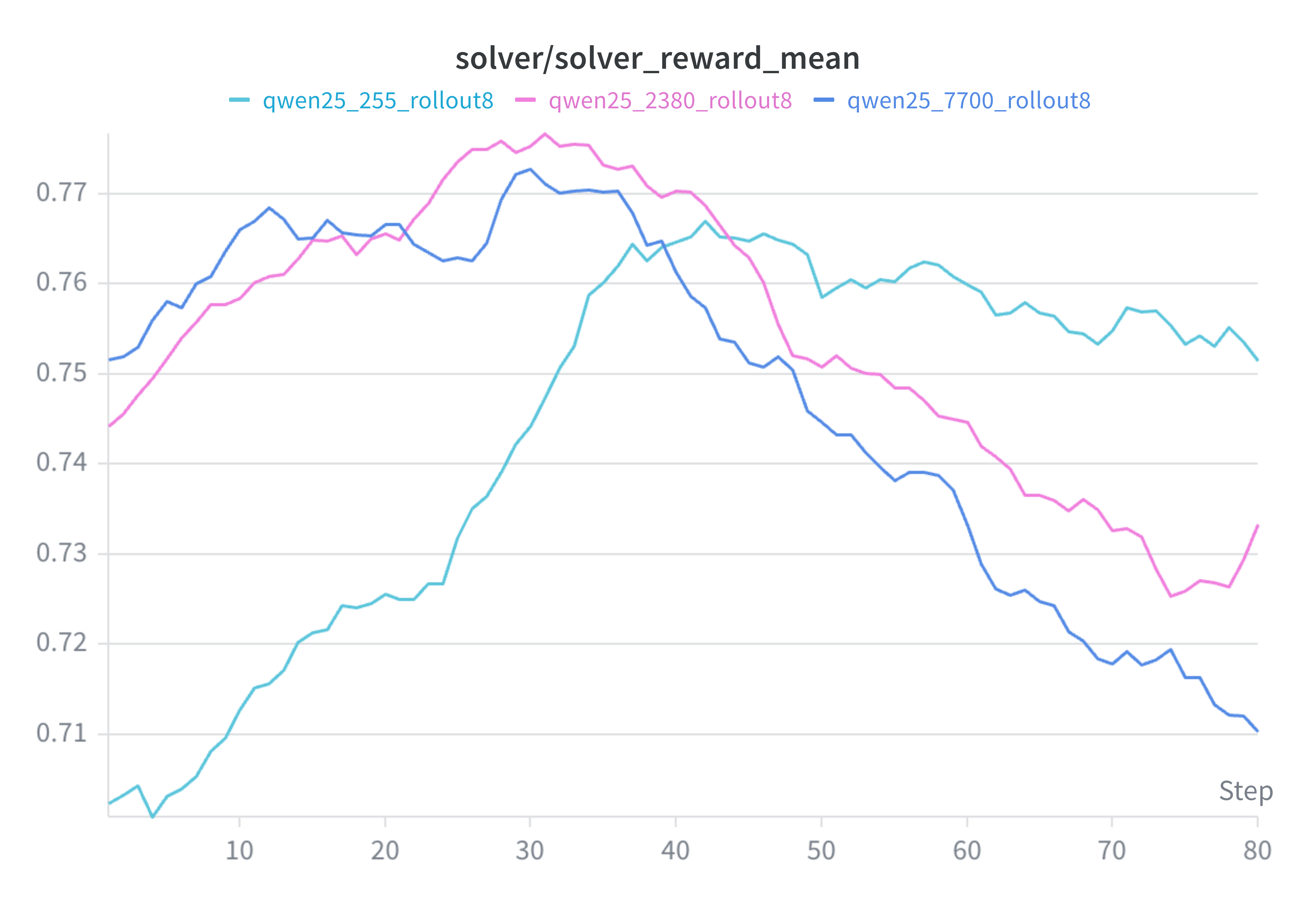}
        \caption{Solver Reward}
    \end{subfigure}\hfill
    \caption{Training Details}
    \label{datasize_ablation}
\end{figure}

\paragraph{Training Data Size}
We apply different training data size in {\ourmethod}. While the co-evolving evaluator successfully stabilizes training on smaller datasets, increasing the data scale accelerates model collapse. The reason is both of the solver and evaluator rely on instinct reward rather than strictly verifiable external ground truth. With limited training data, this exact internal feedback effectively sharpens the model's correct priors and provides necessary dense constraints. However, as the dataset size increases significantly, the solver progressively discovers blind spots in the evaluator and over-optimizes against these proxy rewards. This dynamic inevitably forces the optimization process to saturate, degrading the factual accuracy as the solver exploits the distributional limits of the proxy feedback. As illustrated by Figure~\ref{datasize_ablation}, the solver entropy collapses prematurely at large scales, followed by a classic rise-then-fall trajectory in the training rewards. This observation aligns perfectly with recent theoretical findings on unsupervised reinforcement learning, confirming that intrinsic reward mechanisms eventually amplify self-reinforcing hallucinations when exposed to extensive data \citep{he2026how}.

\paragraph{Case Study}
\input{files/case}
 Figure~\ref{case1} demonstrates the stark contrast in scoring behaviors between static and adaptive evaluations. Under the static rubric, the solver successfully exploits the generic "coherence" metric, securing a near-perfect score of 8 in both scenarios by generating superficially fluent prose. However, this high structural score masks critical omissions in substantive content. When our adaptive framework is applied, it immediately synthesizes query-specific constraints to interrogate the text. In the legal scenario, the dynamic evaluation criteria focus on whether the report grounds its claims in specific statutory prerequisites. Similarly, in the discussion scenario, the metrics expose the report's failure to synthesize competing viewpoints, despite its seemingly thoughtful tone. This dynamic assessment effectively prevents the model from prioritizing elegant phrasing to secure high scores while neglecting factual accuracy or other critical dimensional requirements.
 
\clearpage

\section{Prompts}
\begin{lstlisting}[breaklines=true,label={prompt:drb_vc_prompt},caption={Prompt for citation checking.},captionpos=b]
You will be provided with a reference and some statements. Please determine whether each statement is 'supported', 'unsupported', or 'unknown' with respect to the reference. Please note:
First, assess whether the reference contains any valid content. If the reference contains no valid information, such as a 'page not found' message, then all statements should be considered 'unknown'.
If the reference is valid, for a given statement: if the facts or data it contains can be found entirely or partially within the reference, it is considered 'supported' (data accepts rounding); if all facts and data in the statement cannot be found in the reference, it is considered 'unsupported'.

You should return the result in a JSON list format, where each item in the list contains the statement's index and the judgment result, for example:
[
    {{
        "idx": 1,
        "result": "supported"
    }},
    {{
        "idx": 2,
        "result": "unsupported"
    }}
]

Below are the reference and statements:
<reference>
{reference}
</reference>

<statements>
{statements}
</statements>

Begin the assessment now. Output only the JSON list, without any conversational text or explanations.
\end{lstlisting}

\begin{lstlisting}[breaklines=true,label={prompt:drb_prompt},caption={Prompt for writing quality evaluation},captionpos=b]
<system_role>
You are an experienced research article evaluation expert. Your task is to evaluate a deep research article based on the provided `<task>` across four dimensions: Comprehensiveness, Insight, Instruction Following, and Readability.
</system_role>

<task>
"{task_prompt}"
</task>

<instruction>
Please deeply analyze the task and evaluate the research report based on the following specific criteria for each dimension:

1. Comprehensiveness (Information Coverage)
- Evaluate the breadth, depth, and relevance of information coverage. 
- Ensure the article covers key information areas, perspectives, and depths necessary to achieve comprehensive market or situational analysis.
- Check if it thoroughly examines all relevant factors (e.g., technical, economic, social, geographical) and avoids omitting critical facets of the topic.

2. Insight (Analytical Depth)
- Evaluate the depth, originality, logic, and value of the analysis and conclusions.
- Assess whether the analysis goes beyond a superficial listing of factors to uncover core drivers, subtle mechanisms, and second-order effects.
- Examine if the conclusions and recommendations are logically derived from the analysis, offer unique perspectives, and provide forward-looking or actionable strategic value.

3. Instruction Following (Responsiveness & Relevance)
- Evaluate whether the report accurately, completely, and directly responds to all specific instructions, questions, and core objectives within the `<task>`.
- Check for strict adherence to scope limitations (e.g., specific geography, time, or subject constraints) without significant deviation into irrelevant topics.

4. Readability (Presentation Quality)
- Evaluate the clarity of structure, fluency of language, effectiveness of data presentation, and overall ease of understanding.
- Assess structural logic (clear framework and logical flow), language precision (fluent, grammatically correct, and appropriate use of terminology), and cohesive paragraph transitions.
- Check the clarity and accuracy of data presentation, including the effective use of visualizations (charts, tables) and the highlighting of key findings to reduce reader fatigue.
</instruction>
\end{lstlisting}

\begin{lstlisting}[breaklines=true,label={prompt:meta_harness_prompt},caption={Prompt for meta-harness},captionpos=b]
[System Prompt]
You are an external researcher. Adjust the evaluator harness based on recent training statistics. Do not choose the evaluator's final dimensions or weights directly.
To request search from the evaluator, set require_search=true in process_constraints, and explicitly state in notes that the evaluator should use the search tool by outputting <search>query</search> tags.

[User Prompt]
Task: Generate the next evaluator harness H_t.
Goal: Update evaluator constraints without directly choosing concrete dimensions or weights for the evaluator. You may adjust dimension_constraints and process_constraints and dimension_pool.

## Statistics: 
{Training Statistics JSON}

## Output Format:
{
  "dimension_constraints": {
    "k": number of dimensions,
    "must_include": [list of required dimension names],
    "weight_min": minimum weight for a dimension,
    "weight_max": maximum weight for a dimension
  },
  "process_constraints": {
    "require_gap_analysis": true/false,
    "require_verification": true/false,
    "citation_cross_check": true/false,
    "min_number_of_tool_call": n
  },
  "notes": "additional process compliance instructions"
}
\end{lstlisting}

\begin{lstlisting}[breaklines=true,label={prompt:evaluator_prompt},caption={Prompt for evaluator},captionpos=b]
[System Prompt]
You are a professional evaluator for deep research reports.
You will receive one query and multiple candidate reports generated by the solver.
Your job is to evaluate every report with a search-assisted review process.

You must complete 4 stages:
1. {must_include} is always required as the primary scoring dimension. Select exactly {k-1} additional dimensions from the remaining pool and assign weights, each additional dimension's weight must be within [{Weight Min}, {Weight Max}]
2. Verify key claims and citations in each report using the available search tool.
3. Perform independent research from the original query to identify important gaps not covered by each report.
4. Score each report on every selected dimension using the evidence you found.

## Tool-use protocol:
- When you need external evidence, you can use search tools with search query in this format:
  <search>your search query here</search>
- The content inside <search> must be only the query text. Do not output JSON, function-call syntax, XML attributes, bullet lists, or multiple queries.
- After a search, the system will return retrieved evidence inside <observation>...</observation>.
- You may issue multiple search queries as needed before producing the final XML.

Return the final result in the following XML format ONLY AFTER you finish your search and evaluation:
<dimensions>
  <dim name="[Base Dimension]"/>
  <dim name="[Selected Dimension]" weight="[Weight]"/>
</dimensions>
<report_evaluations>
  <report id="[Report ID]">
    <verification>your verification process here</verification>
    <gap_analysis>your gap analysis here</gap_analysis>
    <scores>
      <score dim="[Base Dimension]" value="[Score]" reason="..."/>
      <score dim="[Selected Dimension]" value="[Score]" reason="..."/>
    </scores>
  </report>
</report_evaluations>
\end{lstlisting}

\begin{lstlisting}[breaklines=true,label={prompt:solver_prompt},caption={Prompt for solver},captionpos=b]
[System Prompt]
You are a professional deep researcher. Solve the user's request by iteratively searching public information and synthesizing evidence.

## Rules:
1. Think inside <think> and </think> before each action and after each new search result.
2. When external evidence is needed, emit exactly one standalone search query. Use the exact format <search>your query here</search> to issue a search.
3. The content inside <search> must be only the query text. Do not output JSON, tool names, XML attributes, bullet lists, or multiple queries in one search block.
4. Prefer natural web search queries: start broad, then narrow to verification, recent developments, entity-specific facts, or missing perspectives.
5. Prefer multiple sequential searches over one overloaded query.
6. Use search to verify important claims, gather evidence, and close information gaps. Stop searching once you can answer confidently.
7. The environment will return search results between <observation> and </observation>. 
8. When you have enough evidence, write the final response only inside <answer> and </answer>.
9. In the final answer, synthesize the evidence, note uncertainty when necessary, and include source mentions or a short Sources section.
\end{lstlisting}

%% file: files/deepresearcheval.tex
\begin{table}[t] 
    \centering
    \small % 使用 small 字号
    \caption{Performance of open-deep-research and gpt-researcher integrated with LLaMA-3.1 and Qwen2.5 on DeepResearchEval. \textbf{Cove}: Breadth, depth, and relevance of coverage.
\textbf{Insight}: Depth, originality, logic, and value of analysis.
\textbf{IF}: Accuracy in meeting all requirements.
\textbf{Clar}: Clarity, fluency, structure, and ease of understanding.
\textbf{Meta}: Query-specific meta-evaluation dimensions generated dynamically for each task.}
    \label{deepresearcheval}
    
    % 极致紧凑化设置
    \renewcommand{\arraystretch}{0.95} % 压缩行高
    \setlength{\tabcolsep}{3.5pt}      % 压缩列间距
    
    \begin{tabular}{l ccccc}
    \toprule
    \textbf{Models and Agents} & \textbf{Cove} & \textbf{Insight} & \textbf{IF.} & \textbf{Clar} & \textbf{Meta} \\
    \midrule
    
    % ================= 分组 1: Open-deep-research =================
    \multicolumn{6}{l}{\textit{\textcolor{gray}{Agent: Open-deep-research (ReAct Paradigm)}}} \\
    
    % LLaMA 组
    \rowcolor{gray!5}
    \hspace{8pt}Llama-3.1-8B-Inst. & 3.7 & 3.2 & 4.5 & 5.1 & 2.6 \\
    \rowcolor{gray!5}
    \hspace{16pt}+ GRPO & 
     1.2 \textcolor{red!70!black}{\scriptsize (-2.5)} & 1.3 \textcolor{red!70!black}{\scriptsize (-1.9)} & 
     1.4 \textcolor{red!70!black}{\scriptsize (-3.1)} & 1.8 \textcolor{red!70!black}{\scriptsize (-3.3)} & 
     1.0 \textcolor{red!70!black}{\scriptsize (-1.6)} \\
    \rowcolor{gray!5}
    \hspace{16pt}+ DPO & 
     4.3  \textcolor{green!50!black}{\scriptsize (+0.6)} &  4.3 \textcolor{green!50!black}{\scriptsize (+1.1)} & 
     4.7  \textcolor{green!50!black}{\scriptsize (+0.2)} & 5.2 \textcolor{green!50!black}{\scriptsize (+0.1)} & 
     3.5 \textcolor{green!50!black}{\scriptsize (+0.9)} \\
    \rowcolor{blue!5}
    \hspace{16pt}+ {\ourmethod} (Ours) & 
     4.3 \textcolor{green!50!black}{\scriptsize (+0.6)} & 
     4.4 \textcolor{green!50!black}{\scriptsize (+1.2)} & 
     4.7 \textcolor{green!50!black}{\scriptsize (+0.2)} & 
     5.2 \textcolor{green!50!black}{\scriptsize (+0.1)} & 
     3.5 \textcolor{green!50!black}{\scriptsize (+0.9)} \\
    
    \addlinespace[1pt] % 仅留 1pt 的极小缝隙区分 LLaMA 和 Qwen
    
    % Qwen 组
    \rowcolor{gray!5}
    \hspace{8pt}Qwen2.5-7B-Inst. & 4.4 & 3.9 & 5.2 & 5.4 & 3.2 \\
    \rowcolor{gray!5}
    \hspace{16pt}+ GRPO & 
     4.4 \textcolor{green!50!black}{\scriptsize (+0.0)} & 4.0   \textcolor{green!50!black}{\scriptsize (+0.1)} & 
     5.2 \textcolor{green!50!black}{\scriptsize (+0.0)} & 5.4 \textcolor{green!50!black}{\scriptsize (+0.0)} & 
     3.3 \textcolor{green!50!black}{\scriptsize (+0.1)} \\
    \rowcolor{gray!5}
    \hspace{16pt}+ DPO & 
    4.6  \textcolor{green!50!black}{\scriptsize (+0.2)} & 4.1 \textcolor{green!50!black}{\scriptsize (+0.2)} & 
      5.6 \textcolor{green!50!black}{\scriptsize (+0.4)} & 5.8 \textcolor{green!50!black}{\scriptsize (+0.4)} & 
     3.4 \textcolor{green!50!black}{\scriptsize (+0.2)} \\
    \rowcolor{blue!5}
    \hspace{16pt}+ {\ourmethod} (Ours) & 
     6.3 \textcolor{green!50!black}{\scriptsize (+1.9)} & 
     5.9 \textcolor{green!50!black}{\scriptsize (+2.0)} & 
     7.0 \textcolor{green!50!black}{\scriptsize (+1.8)} & 
     6.6 \textcolor{green!50!black}{\scriptsize (+1.2)} & 
     5.0 \textcolor{green!50!black}{\scriptsize (+1.8)} \\
    
    \midrule
    
    % ================= 分组 2: gpt-researcher =================
    \multicolumn{6}{l}{\textit{\textcolor{gray}{Agent: gpt-researcher (Plan-and-Execute Paradigm)}}} \\
    
    % LLaMA 组
    \rowcolor{gray!5}
    \hspace{8pt}Llama-3.1-8B-Inst. & 3.9 & 3.5 & 4.8 & 5.6 & 2.8\\
    \rowcolor{gray!5}
    \hspace{16pt}+ GRPO & 
    2.5 \textcolor{red!70!black}{\scriptsize (-1.4)} & 2.8 \textcolor{red!70!black}{\scriptsize (-0.7)} & 
     2.7 \textcolor{red!70!black}{\scriptsize (-2.1)} & 3.7 \textcolor{red!70!black}{\scriptsize (-1.9)} & 
     2.0 \textcolor{red!70!black}{\scriptsize (-0.8)} \\
    \rowcolor{gray!5}
    \hspace{16pt}+ DPO & 
    2.4              \textcolor{red!70!black}{\scriptsize (-1.5)} &  2.8 \textcolor{red!70!black}{\scriptsize (-0.7)} & 
      2.6 \textcolor{red!70!black}{\scriptsize (-2.2)} & 3.5 \textcolor{red!70!black}{\scriptsize (-2.1)} & 
    2.0 \textcolor{red!70!black}{\scriptsize (-0.8)} \\
    \rowcolor{blue!5}
    \hspace{16pt}+ {\ourmethod} (Ours) & 
     5.9 \textcolor{green!50!black}{\scriptsize (+2.0)} & 5.6 \textcolor{green!50!black}{\scriptsize (+2.1)} & 
     6.9 \textcolor{green!50!black}{\scriptsize (+2.1)} & 6.9 \textcolor{green!50!black}{\scriptsize (+1.3)} & 
     4.6 \textcolor{green!50!black}{\scriptsize (+1.8)} \\
    
    \addlinespace[1pt]
    
    % Qwen 组
    \rowcolor{gray!5}
    \hspace{8pt}Qwen2.5-7B-Inst. & 4.5 & 4.1 & 5.7 & 5.9 & 3.4 \\
    \rowcolor{gray!5}
    \hspace{16pt}+ GRPO & 
    4.6 \textcolor{green!50!black}{\scriptsize (+0.1)} & 4.1 \textcolor{green!50!black}{\scriptsize (+0.0)} & 
    5.6 \textcolor{red!70!black}{\scriptsize (-0.1)} & 5.8 \textcolor{red!70!black}{\scriptsize (-0.1)} & 
    3.4 \textcolor{green!50!black}{\scriptsize (+0.0)} \\
    \rowcolor{gray!5}
    \hspace{16pt}+ DPO & 
     6.1 \textcolor{green!50!black}{\scriptsize (+1.6)} & 5.8 \textcolor{green!50!black}{\scriptsize (+1.7)} & 
     6.8 \textcolor{green!50!black}{\scriptsize (+1.1)} & 6.5 \textcolor{green!50!black}{\scriptsize (+0.6)} & 
     4.8 \textcolor{green!50!black}{\scriptsize (+1.4)} \\
    \rowcolor{blue!5}
    \hspace{16pt}+ {\ourmethod} (Ours) & 
     6.5 \textcolor{green!50!black}{\scriptsize (+2.0)} & 6.0 \textcolor{green!50!black}{\scriptsize (+1.9)} & 
     7.5 \textcolor{green!50!black}{\scriptsize (+1.8)} & 7.0 \textcolor{green!50!black}{\scriptsize (+1.1)} & 
     5.1 \textcolor{green!50!black}{\scriptsize (+1.7)} \\
    
    \bottomrule
    \end{tabular}
\end{table}

%% file: files/case.tex
\begin{figure}[t]
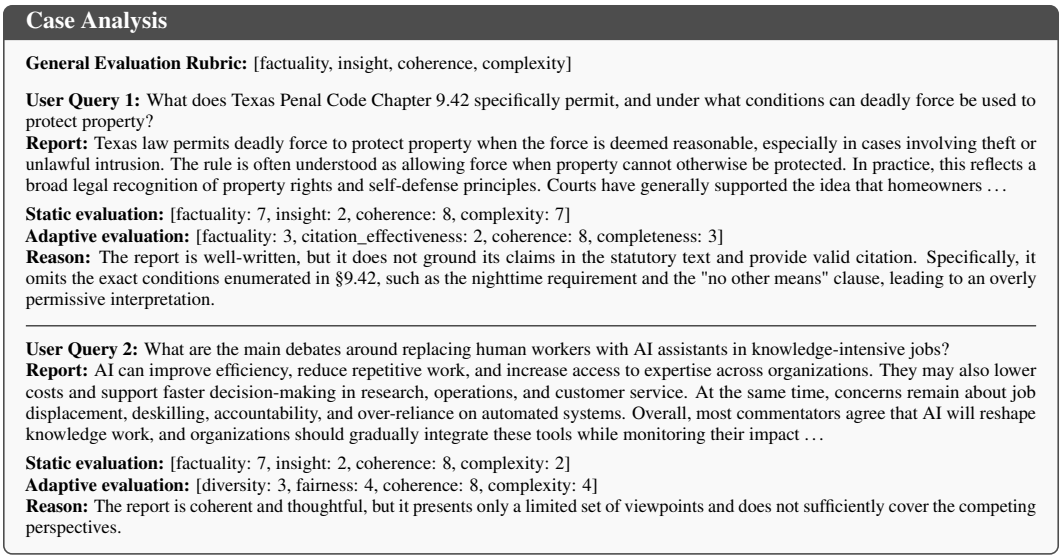

    \centering
\begin{tcolorbox}[
  enhanced,
  width=\linewidth,
  title={\small \textbf{Case Analysis}},
  coltitle=white,
  colbacktitle=black!70,
  colback=gray!5,
  colframe=black!70,
  boxrule=0.6pt,
  arc=1mm,
  left=6pt, right=6pt, top=4pt, bottom=4pt,
  boxsep=2pt,
  fontupper=\scriptsize,
  before skip=4pt,
  after skip=4pt
]
\textbf{General Evaluation Rubric:} [factuality, insight, coherence, complexity]

\medskip
\textbf{User Query 1:} What does Texas Penal Code Chapter 9.42 specifically permit, and under what conditions can deadly force be used to protect property?

\textbf{Report:} Texas law permits deadly force to protect property when the force is deemed reasonable, especially in cases involving theft or unlawful intrusion. The rule is often understood as allowing force when property cannot otherwise be protected. In practice, this reflects a broad legal recognition of property rights and self-defense principles. Courts have generally supported the idea that homeowners \ldots

\smallskip
\textbf{Static evaluation:} [factuality: 7, insight: 2, coherence: 8, complexity: 7]

\textbf{Adaptive evaluation:} [factuality: 3, citation\_effectiveness: 2, coherence: 8, completeness: 3]

\textbf{Reason:} The report is well-written, but it does not ground its claims in the statutory text and provide valid citation. Specifically, it omits the exact conditions enumerated in \S 9.42, such as the nighttime requirement and the "no other means" clause, leading to an overly permissive interpretation.

\medskip
\hrule
\medskip

\textbf{User Query 2:} What are the main debates around replacing human workers with AI assistants in knowledge-intensive jobs?

\textbf{Report:} AI can improve efficiency, reduce repetitive work, and increase access to expertise across organizations. They may also lower costs and support faster decision-making in research, operations, and customer service. At the same time, concerns remain about job displacement, deskilling, accountability, and over-reliance on automated systems. Overall, most commentators agree that AI will reshape knowledge work, and organizations should gradually integrate these tools while monitoring their impact \ldots

\smallskip
\textbf{Static evaluation:} [factuality: 7, insight: 2, coherence: 8, complexity: 2]

\textbf{Adaptive evaluation:} [diversity: 3, fairness: 4, coherence: 8, complexity: 4]

\textbf{Reason:} The report is coherent and thoughtful, but it presents only a limited set of viewpoints and does not sufficiently cover the competing perspectives.
\end{tcolorbox}
    \caption{Adaptive evaluation vs static evaluation of different queries.}
    \label{case1}
    \vspace{-20pt}
\end{figure}

%% file: custom.bib
@article{williams1992simple,
  title={Simple statistical gradient-following algorithms for connectionist reinforcement learning},
  author={Williams, Ronald J},
  journal={Machine learning},
  volume={8},
  number={3},
  pages={229--256},
  year={1992},
  publisher={Springer}
}

@article{zhang2024survey,
  title={A survey on self-play methods in reinforcement learning},
  author={Zhang, Ruize and Xu, Zelai and Ma, Chengdong and Yu, Chao and Tu, Wei-Wei and Tang, Wenhao and Huang, Shiyu and Ye, Deheng and Ding, Wenbo and Yang, Yaodong and others},
  journal={arXiv preprint arXiv:2408.01072},
  year={2024}
}

@inproceedings{
zhao2025absolute,
title={Absolute Zero: Reinforced Self-play Reasoning with Zero Data},
author={Andrew Zhao and Yiran Wu and Yang Yue and Tong Wu and Quentin Xu and Yang Yue and Matthieu Lin and Shenzhi Wang and Qingyun Wu and Zilong Zheng and Gao Huang},
booktitle={The Thirty-ninth Annual Conference on Neural Information Processing Systems},
year={2025},
url={https://openreview.net/forum?id=neZSGqhxDa}
}

@inproceedings{
huang2025r,
title={R-Zero: Self-Evolving Reasoning {LLM} from Zero Data},
author={Chengsong Huang and Wenhao Yu and Xiaoyang Wang and Hongming Zhang and Zongxia Li and Ruosen Li and Jiaxin Huang and Haitao Mi and Dong Yu},
booktitle={The Fourteenth International Conference on Learning Representations},
year={2026},
url={https://openreview.net/forum?id=96apU6YzSO}
}

@inproceedings{
liu2025spiral,
title={{SPIRAL}: Self-Play on Zero-Sum Games Incentivizes Reasoning via Multi-Agent Multi-Turn Reinforcement Learning},
author={Bo Liu and Simon Yu and Zichen Liu and Leon Guertler and Penghui Qi and Daniel Balcells and Mickel Liu and Cheston Tan and Weiyan Shi and Min Lin and Wee Sun Lee and Natasha Jaques},
booktitle={The Fourteenth International Conference on Learning Representations},
year={2026},
url={https://openreview.net/forum?id=7Yayy5fNLg}
}

@inproceedings{
lu2025search,
title={Search Self-Play: Pushing the Frontier of Agent Capability without Supervision},
author={Hongliang Lu and Yuhang Wen and Pengyu Cheng and Ruijin Ding and Haotian Xu and Jiaqi Guo and Chutian Wang and Haonan Chen and xiaoxi jiang and guanjunjiang},
booktitle={The Fourteenth International Conference on Learning Representations},
year={2026},
url={https://openreview.net/forum?id=ZmGirmNJqE}
}

@article{yue2026dr,
  title={Dr. Zero: Self-Evolving Search Agents without Training Data},
  author={Yue, Zhenrui and Upasani, Kartikeya and Yang, Xianjun and Ge, Suyu and Nie, Shaoliang and Mao, Yuning and Liu, Zhe and Wang, Dong},
  journal={arXiv preprint arXiv:2601.07055},
  year={2026}
}

@inproceedings{
cheng2025sparselfplaytreesearchrefinement,
title={{SP}aR: Self-Play with Tree-Search Refinement to Improve Instruction-Following in Large Language Models},
author={Jiale Cheng and Xiao Liu and Cunxiang Wang and Xiaotao Gu and Yida Lu and Dan Zhang and Yuxiao Dong and Jie Tang and Hongning Wang and Minlie Huang},
booktitle={The Thirteenth International Conference on Learning Representations},
year={2025},
url={https://openreview.net/forum?id=9chRqsPOGL}
}

@inproceedings{
fang2026serl,
title={Se{RL}: Self-play Reinforcement Learning for Large Language Models with Limited Data},
author={Wenkai Fang and Shunyu Liu and Yang Zhou and Kongcheng Zhang and Tongya Zheng and Kaixuan Chen and Mingli Song and Dacheng Tao},
booktitle={The Thirty-ninth Annual Conference on Neural Information Processing Systems},
year={2026},
url={https://openreview.net/forum?id=ZF93vyH9He}
}

@inproceedings{
li2026webthinker,
title={WebThinker: Empowering Large Reasoning Models with Deep Research Capability},
author={Xiaoxi Li and Jiajie Jin and Guanting Dong and Hongjin Qian and Yongkang Wu and Ji-Rong Wen and Yutao Zhu and Zhicheng Dou},
booktitle={The Thirty-ninth Annual Conference on Neural Information Processing Systems},
year={2026},
url={https://openreview.net/forum?id=7LKKHBAMzH}
}

@inproceedings{zhang-etal-2025-evolvesearch,
    title = "{E}volve{S}earch: An Iterative Self-Evolving Search Agent",
    author = "Zhang, Ding-Chu  and
      Zhao, Yida  and
      Wu, Jialong  and
      Zhang, Liwen  and
      Li, Baixuan  and
      Yin, Wenbiao  and
      Jiang, Yong  and
      Li, Yu-Feng  and
      Tu, Kewei  and
      Xie, Pengjun  and
      Huang, Fei",
    editor = "Christodoulopoulos, Christos  and
      Chakraborty, Tanmoy  and
      Rose, Carolyn  and
      Peng, Violet",
    booktitle = "Proceedings of the 2025 Conference on Empirical Methods in Natural Language Processing",
    month = nov,
    year = "2025",
    address = "Suzhou, China",
    publisher = "Association for Computational Linguistics",
    url = "https://aclanthology.org/2025.emnlp-main.663/",
    doi = "10.18653/v1/2025.emnlp-main.663",
    pages = "13123--13136",
    ISBN = "979-8-89176-332-6"
}

@inproceedings{
shi2026iterative,
title={Iterative Self-Incentivization Empowers Large Language Models as Agentic Searchers},
author={Zhengliang Shi and Lingyong Yan and Dawei Yin and Suzan Verberne and Maarten de Rijke and Zhaochun Ren},
booktitle={The Thirty-ninth Annual Conference on Neural Information Processing Systems},
year={2026},
url={https://openreview.net/forum?id=s9NkfkUuEr}
}

@inproceedings{
xu2026acesearcher,
title={AceSearcher: Bootstrapping Reasoning and Search for {LLM}s via Reinforced Self-Play},
author={Ran Xu and Yuchen Zhuang and Zihan Dong and Ruiyu Wang and Yue Yu and Joyce C. Ho and Linjun Zhang and Haoyu Wang and Wenqi Shi and Carl Yang},
booktitle={The Thirty-ninth Annual Conference on Neural Information Processing Systems},
year={2026},
url={https://openreview.net/forum?id=jSgCM0uZn3}
}

@inproceedings{
zhao2026learning,
title={Learning to Reason without External Rewards},
author={Xuandong Zhao and Zhewei Kang and Aosong Feng and Sergey Levine and Dawn Song},
booktitle={The Fourteenth International Conference on Learning Representations},
year={2026},
url={https://openreview.net/forum?id=OU9nFEYR2M}
}

@inproceedings{
tang2026beyond,
title={Beyond Verifiable Rewards: Scaling Reinforcement Learning in Language Models to Unverifiable Data},
author={Yunhao Tang and Sid Wang and Lovish Madaan and Remi Munos},
booktitle={The Thirty-ninth Annual Conference on Neural Information Processing Systems},
year={2026},
url={https://openreview.net/forum?id=pc6M9h3T9m}
}

@inproceedings{
jin2025searchr,
title={Search-R1: Training {LLM}s to Reason and Leverage Search Engines with Reinforcement Learning},
author={Bowen Jin and Hansi Zeng and Zhenrui Yue and Jinsung Yoon and Sercan O Arik and Dong Wang and Hamed Zamani and Jiawei Han},
booktitle={Second Conference on Language Modeling},
year={2025},
url={https://openreview.net/forum?id=Rwhi91ideu}
}

@misc{song2025r1searcherincentivizingsearchcapability,
      title={R1-Searcher: Incentivizing the Search Capability in LLMs via Reinforcement Learning}, 
      author={Huatong Song and Jinhao Jiang and Yingqian Min and Jie Chen and Zhipeng Chen and Wayne Xin Zhao and Lei Fang and Ji-Rong Wen},
      year={2025},
      eprint={2503.05592},
      archivePrefix={arXiv},
      primaryClass={cs.AI},
      url={https://arxiv.org/abs/2503.05592}, 
}

@inproceedings{li-etal-2025-search,
    title = "Search-o1: Agentic Search-Enhanced Large Reasoning Models",
    author = "Li, Xiaoxi  and
      Dong, Guanting  and
      Jin, Jiajie  and
      Zhang, Yuyao  and
      Zhou, Yujia  and
      Zhu, Yutao  and
      Zhang, Peitian  and
      Dou, Zhicheng",
    editor = "Christodoulopoulos, Christos  and
      Chakraborty, Tanmoy  and
      Rose, Carolyn  and
      Peng, Violet",
    booktitle = "Proceedings of the 2025 Conference on Empirical Methods in Natural Language Processing",
    month = nov,
    year = "2025",
    address = "Suzhou, China",
    publisher = "Association for Computational Linguistics",
    url = "https://aclanthology.org/2025.emnlp-main.276/",
    doi = "10.18653/v1/2025.emnlp-main.276",
    pages = "5420--5438",
    ISBN = "979-8-89176-332-6"
}

@inproceedings{zheng-etal-2025-deepresearcher,
    title = "{D}eep{R}esearcher: Scaling Deep Research via Reinforcement Learning in Real-world Environments",
    author = "Zheng, Yuxiang  and
      Fu, Dayuan  and
      Hu, Xiangkun  and
      Cai, Xiaojie  and
      Ye, Lyumanshan  and
      Lu, Pengrui  and
      Liu, Pengfei",
    editor = "Christodoulopoulos, Christos  and
      Chakraborty, Tanmoy  and
      Rose, Carolyn  and
      Peng, Violet",
    booktitle = "Proceedings of the 2025 Conference on Empirical Methods in Natural Language Processing",
    month = nov,
    year = "2025",
    address = "Suzhou, China",
    publisher = "Association for Computational Linguistics",
    url = "https://aclanthology.org/2025.emnlp-main.22/",
    doi = "10.18653/v1/2025.emnlp-main.22",
    pages = "414--431",
    ISBN = "979-8-89176-332-6"
}

@inproceedings{sun-etal-2025-simpledeepsearcher,
    title = "{S}imple{D}eep{S}earcher: Deep Information Seeking via Web-Powered Reasoning Trajectory Synthesis",
    author = "Sun, Shuang  and
      Song, Huatong  and
      Wang, Yuhao  and
      Ren, Ruiyang  and
      Jiang, Jinhao  and
      Zhang, Junjie  and
      Bai, Fei  and
      Deng, Jia  and
      Zhao, Wayne Xin  and
      Liu, Zheng  and
      Fang, Lei  and
      Wang, Zhongyuan  and
      Wen, Ji-Rong",
    editor = "Christodoulopoulos, Christos  and
      Chakraborty, Tanmoy  and
      Rose, Carolyn  and
      Peng, Violet",
    booktitle = "Findings of the Association for Computational Linguistics: EMNLP 2025",
    month = nov,
    year = "2025",
    address = "Suzhou, China",
    publisher = "Association for Computational Linguistics",
    url = "https://aclanthology.org/2025.findings-emnlp.739/",
    doi = "10.18653/v1/2025.findings-emnlp.739",
    pages = "13705--13720",
    ISBN = "979-8-89176-335-7"
}

@misc{li2026agentcpmreportinterleavingdraftingdeepening,
      title={AgentCPM-Report: Interleaving Drafting and Deepening for Open-Ended Deep Research}, 
      author={Yishan Li and Wentong Chen and Yukun Yan and Mingwei Li and Sen Mei and Xiaorong Wang and Kunpeng Liu and Xin Cong and Shuo Wang and Zhong Zhang and Yaxi Lu and Zhenghao Liu and Yankai Lin and Zhiyuan Liu and Maosong Sun},
      year={2026},
      eprint={2602.06540},
      archivePrefix={arXiv},
      primaryClass={cs.AI},
      url={https://arxiv.org/abs/2602.06540}, 
}

@misc{hu2025stepdeepresearchtechnicalreport,
      title={Step-DeepResearch Technical Report}, 
      author={Chen Hu and Haikuo Du and Heng Wang and Lin Lin and Mingrui Chen and Peng Liu and Ruihang Miao and Tianchi Yue and Wang You and Wei Ji and Wei Yuan and Wenjin Deng and Xiaojian Yuan and Xiaoyun Zhang and Xiangyu Liu and Xikai Liu and Yanming Xu and Yicheng Cao and Yifei Zhang and Yongyao Wang and Yubo Shu and Yurong Zhang and Yuxiang Zhang and Zheng Gong and Zhichao Chang and Binyan Li and Dan Ma and Furong Jia and Hongyuan Wang and Jiayu Liu and Jing Bai and Junlan Liu and Manjiao Liu and Na Wang and Qiuping Wu and Qinxin Du and Shiwei Li and Wen Sun and Yifeng Gong and Yonglin Chen and Yuling Zhao and Yuxuan Lin and Ziqi Ren and Zixuan Wang and Aihu Zhang and Brian Li and Buyun Ma and Kang An and Li Xie and Mingliang Li and Pan Li and Shidong Yang and Xi Chen and Xiaojia Liu and Yuchu Luo and Yuan Song and YuanHao Ding and Yuanwei Liang and Zexi Li and Zhaoning Zhang and Zixin Zhang and Binxing Jiao and Daxin Jiang and Jiansheng Chen and Jing Li and Xiangyu Zhang and Yibo Zhu},
      year={2025},
      eprint={2512.20491},
      archivePrefix={arXiv},
      primaryClass={cs.CL},
      url={https://arxiv.org/abs/2512.20491}, 
}

@inproceedings{jiang-etal-2024-unknown,
    title = "Into the Unknown Unknowns: Engaged Human Learning through Participation in Language Model Agent Conversations",
    author = "Jiang, Yucheng  and
      Shao, Yijia  and
      Ma, Dekun  and
      Semnani, Sina  and
      Lam, Monica",
    editor = "Al-Onaizan, Yaser  and
      Bansal, Mohit  and
      Chen, Yun-Nung",
    booktitle = "Proceedings of the 2024 Conference on Empirical Methods in Natural Language Processing",
    month = nov,
    year = "2024",
    address = "Miami, Florida, USA",
    publisher = "Association for Computational Linguistics",
    url = "https://aclanthology.org/2024.emnlp-main.554/",
    doi = "10.18653/v1/2024.emnlp-main.554",
    pages = "9917--9955"
}

@inproceedings{shao-etal-2024-assisting,
    title = "Assisting in Writing {W}ikipedia-like Articles From Scratch with Large Language Models",
    author = "Shao, Yijia  and
      Jiang, Yucheng  and
      Kanell, Theodore  and
      Xu, Peter  and
      Khattab, Omar  and
      Lam, Monica",
    editor = "Duh, Kevin  and
      Gomez, Helena  and
      Bethard, Steven",
    booktitle = "Proceedings of the 2024 Conference of the North American Chapter of the Association for Computational Linguistics: Human Language Technologies (Volume 1: Long Papers)",
    month = jun,
    year = "2024",
    address = "Mexico City, Mexico",
    publisher = "Association for Computational Linguistics",
    url = "https://aclanthology.org/2024.naacl-long.347/",
    doi = "10.18653/v1/2024.naacl-long.347",
    pages = "6252--6278"
}

@inproceedings{gptresearcher,
    title = "gpt-researcher",
    url = "https://docs.gptr.dev/docs/gpt-researcher/getting-started/introduction"
}

@misc{chen2026agentcpmexplorerealizinglonghorizondeep,
      title={AgentCPM-Explore: Realizing Long-Horizon Deep Exploration for Edge-Scale Agents}, 
      author={Haotian Chen and Xin Cong and Shengda Fan and Yuyang Fu and Ziqin Gong and Yaxi Lu and Yishan Li and Boye Niu and Chengjun Pan and Zijun Song and Huadong Wang and Yesai Wu and Yueying Wu and Zihao Xie and Yukun Yan and Zhong Zhang and Yankai Lin and Zhiyuan Liu and Maosong Sun},
      year={2026},
      eprint={2602.06485},
      archivePrefix={arXiv},
      primaryClass={cs.AI},
      url={https://arxiv.org/abs/2602.06485}, 
}

@article{zheng2023judging,
  title={Judging llm-as-a-judge with mt-bench and chatbot arena},
  author={Zheng, Lianmin and Chiang, Wei-Lin and Sheng, Ying and Zhuang, Siyuan and Wu, Zhanghao and Zhuang, Yonghao and Lin, Zi and Li, Zhuohan and Li, Dacheng and Xing, Eric and others},
  journal={Advances in neural information processing systems},
  volume={36},
  pages={46595--46623},
  year={2023}
}

@inproceedings{
yang2025trajectoryllm,
title={Trajectory-{LLM}: A Language-based Data Generator for Trajectory Prediction in Autonomous Driving},
author={Kairui Yang and Zihao Guo and Gengjie Lin and Haotian Dong and Zhao Huang and Yipeng Wu and Die Zuo and Jibin Peng and Ziyuan Zhong and Xin WANG and Qing Guo and Xiaosong Jia and Junchi Yan and Di Lin},
booktitle={The Thirteenth International Conference on Learning Representations},
year={2025},
url={https://openreview.net/forum?id=UapxTvxB3N}
}

@article{dong2025survey,
  title={A survey on code generation with llm-based agents},
  author={Dong, Yihong and Jiang, Xue and Qian, Jiaru and Wang, Tian and Zhang, Kechi and Jin, Zhi and Li, Ge},
  journal={arXiv preprint arXiv:2508.00083},
  year={2025}
}

@article{seedance2026seedance,
  title={Seedance 2.0: Advancing Video Generation for World Complexity},
  author={Seedance, Team and Chen, De and Chen, Liyang and Chen, Xin and Chen, Ying and Chen, Zhuo and Chen, Zhuowei and Cheng, Feng and Cheng, Tianheng and Cheng, Yufeng and others},
  journal={arXiv preprint arXiv:2604.14148},
  year={2026}
}

@misc{ding2026adarubrictaskadaptiverubricsllm,
      title={AdaRubric: Task-Adaptive Rubrics for LLM Agent Evaluation}, 
      author={Liang Ding},
      year={2026},
      eprint={2603.21362},
      archivePrefix={arXiv},
      primaryClass={cs.AI},
      url={https://arxiv.org/abs/2603.21362}, 
}

@inproceedings{
kim2024prometheus,
title={Prometheus: Inducing Fine-Grained Evaluation Capability in Language Models},
author={Seungone Kim and Jamin Shin and Yejin Cho and Joel Jang and Shayne Longpre and Hwaran Lee and Sangdoo Yun and Seongjin Shin and Sungdong Kim and James Thorne and Minjoon Seo},
booktitle={The Twelfth International Conference on Learning Representations},
year={2024},
url={https://openreview.net/forum?id=8euJaTveKw}
}

@misc{zhang2026lengthscalingsynergizingbreadth,
      title={Beyond Length Scaling: Synergizing Breadth and Depth for Generative Reward Models}, 
      author={Qiyuan Zhang and Yufei Wang and Tianhe Wu and Can Xu and Qingfeng Sun and Kai Zheng and Xue Liu and Chen Ma},
      year={2026},
      eprint={2603.01571},
      archivePrefix={arXiv},
      primaryClass={cs.AI},
      url={https://arxiv.org/abs/2603.01571}, 
}

@inproceedings{
du2026deepresearch,
title={DeepResearch Bench: A Comprehensive Benchmark for Deep Research Agents},
author={Mingxuan Du and Benfeng Xu and Chiwei Zhu and Licheng Zhang and Xiaorui Wang and Zhendong Mao},
booktitle={The Fourteenth International Conference on Learning Representations},
year={2026},
url={https://openreview.net/forum?id=hQ0K2Hhq7H}
}

@misc{lei2026writingrladvancinglongformwriting,
      title={Writing-RL: Advancing Long-form Writing via Adaptive Curriculum Reinforcement Learning}, 
      author={Xuanyu Lei and Chenliang Li and Yuning Wu and Kaiming Liu and Weizhou Shen and Peng Li and Ming Yan and Fei Huang and Ya-Qin Zhang and Yang Liu},
      year={2026},
      eprint={2506.05760},
      archivePrefix={arXiv},
      primaryClass={cs.CL},
      url={https://arxiv.org/abs/2506.05760}, 
}

@misc{gao2026drarenaautomatedevaluationframework,
      title={DR-Arena: an Automated Evaluation Framework for Deep Research Agents}, 
      author={Yiwen Gao and Ruochen Zhao and Yang Deng and Wenxuan Zhang},
      year={2026},
      eprint={2601.10504},
      archivePrefix={arXiv},
      primaryClass={cs.CL},
      url={https://arxiv.org/abs/2601.10504}, 
}

@misc{lee2026metaharnessendtoendoptimizationmodel,
      title={Meta-Harness: End-to-End Optimization of Model Harnesses}, 
      author={Yoonho Lee and Roshen Nair and Qizheng Zhang and Kangwook Lee and Omar Khattab and Chelsea Finn},
      year={2026},
      eprint={2603.28052},
      archivePrefix={arXiv},
      primaryClass={cs.AI},
      url={https://arxiv.org/abs/2603.28052}, 
}

@inproceedings{Sheng_2025, series={EuroSys ’25},
   title={HybridFlow: A Flexible and Efficient RLHF Framework},
   url={http://dx.doi.org/10.1145/3689031.3696075},
   DOI={10.1145/3689031.3696075},
   booktitle={Proceedings of the Twentieth European Conference on Computer Systems},
   publisher={ACM},
   author={Sheng, Guangming and Zhang, Chi and Ye, Zilingfeng and Wu, Xibin and Zhang, Wang and Zhang, Ru and Peng, Yanghua and Lin, Haibin and Wu, Chuan},
   year={2025},
   month=Mar, pages={1279–1297},
   collection={EuroSys ’25} }

@article{DBLP:journals/corr/abs-2402-03300,
  publtype={informal},
  author={Zhihong Shao and Peiyi Wang and Qihao Zhu and Runxin Xu and Junxiao Song and Mingchuan Zhang and Y. K. Li and Y. Wu and Daya Guo},
  title={DeepSeekMath: Pushing the Limits of Mathematical Reasoning in Open Language Models},
  year={2024},
  cdate={1704067200000},
  journal={CoRR},
  volume={abs/2402.03300},
  url={https://doi.org/10.48550/arXiv.2402.03300}
}

@misc{gpt52,
    year=2025,
    title={Introduction GPT-5.2},
  url={https://openai.com/zh-Hans-CN/index/introducing-gpt-5-2/}
}

@misc{qwen2025qwen25technicalreport,
      title={Qwen2.5 Technical Report}, 
      author={Qwen and : and An Yang and Baosong Yang and Beichen Zhang and Binyuan Hui and Bo Zheng and Bowen Yu and Chengyuan Li and Dayiheng Liu and Fei Huang and Haoran Wei and Huan Lin and Jian Yang and Jianhong Tu and Jianwei Zhang and Jianxin Yang and Jiaxi Yang and Jingren Zhou and Junyang Lin and Kai Dang and Keming Lu and Keqin Bao and Kexin Yang and Le Yu and Mei Li and Mingfeng Xue and Pei Zhang and Qin Zhu and Rui Men and Runji Lin and Tianhao Li and Tianyi Tang and Tingyu Xia and Xingzhang Ren and Xuancheng Ren and Yang Fan and Yang Su and Yichang Zhang and Yu Wan and Yuqiong Liu and Zeyu Cui and Zhenru Zhang and Zihan Qiu},
      year={2025},
      eprint={2412.15115},
      archivePrefix={arXiv},
      primaryClass={cs.CL},
      url={https://arxiv.org/abs/2412.15115}, 
}

@misc{grattafiori2024llama3herdmodels,
      title={The Llama 3 Herd of Models}, 
      author={Aaron Grattafiori and Abhimanyu Dubey and Abhinav Jauhri and Abhinav Pandey and Abhishek Kadian and Ahmad Al-Dahle and Aiesha Letman and Akhil Mathur and Alan Schelten and Alex Vaughan and Amy Yang and Angela Fan and others},
      year={2024},
      eprint={2407.21783},
      archivePrefix={arXiv},
      primaryClass={cs.AI},
      url={https://arxiv.org/abs/2407.21783}, 
}

@inproceedings{opendeepresearch,
    title = "open-deep-research",
    url = "https://github.com/langchain-ai/open_deep_research"
}

@Article{app16052546,
AUTHOR = {Tuohetiyaer, Yeerpan and Zhu, Yuye and Hu, Yan and Lu, Siyuan and Wang, Zhongfeng},
TITLE = {Deep-Research Eval: An Automated Framework for Assessing Quality and Reliability in Long-Form Reports},
JOURNAL = {Applied Sciences},
VOLUME = {16},
YEAR = {2026},
NUMBER = {5},
ARTICLE-NUMBER = {2546},
URL = {https://www.mdpi.com/2076-3417/16/5/2546},
ISSN = {2076-3417},
DOI = {10.3390/app16052546}
}

@misc{yang2025qwen3technicalreport,
      title={Qwen3 Technical Report}, 
      author={An Yang and Anfeng Li and Baosong Yang and Beichen Zhang and Binyuan Hui and Bo Zheng and Bowen Yu and Chang Gao and Chengen Huang and Chenxu Lv and Chujie Zheng and Dayiheng Liu and Fan Zhou and Fei Huang and Feng Hu and Hao Ge and Haoran Wei and Huan Lin and Jialong Tang and Jian Yang and Jianhong Tu and Jianwei Zhang and Jianxin Yang and Jiaxi Yang and Jing Zhou and Jingren Zhou and Junyang Lin and Kai Dang and Keqin Bao and Kexin Yang and Le Yu and Lianghao Deng and Mei Li and Mingfeng Xue and Mingze Li and Pei Zhang and Peng Wang and Qin Zhu and Rui Men and Ruize Gao and Shixuan Liu and Shuang Luo and Tianhao Li and Tianyi Tang and Wenbiao Yin and Xingzhang Ren and Xinyu Wang and Xinyu Zhang and Xuancheng Ren and Yang Fan and Yang Su and Yichang Zhang and Yinger Zhang and Yu Wan and Yuqiong Liu and Zekun Wang and Zeyu Cui and Zhenru Zhang and Zhipeng Zhou and Zihan Qiu},
      year={2025},
      eprint={2505.09388},
      archivePrefix={arXiv},
      primaryClass={cs.CL},
      url={https://arxiv.org/abs/2505.09388}, 
}

@misc{5team2025glm45agenticreasoningcoding,
      title={GLM-4.5: Agentic, Reasoning, and Coding (ARC) Foundation Models}, 
      author={ 5 Team and Aohan Zeng and Xin Lv and Qinkai Zheng and Zhenyu Hou and Bin Chen and Chengxing Xie and Cunxiang Wang and Da Yin and Hao Zeng and Jiajie Zhang and Kedong Wang and Lucen Zhong and Mingdao Liu and others},
      year={2025},
      eprint={2508.06471},
      archivePrefix={arXiv},
      primaryClass={cs.CL},
      url={https://arxiv.org/abs/2508.06471}, 
}

@misc{deepseekai2025deepseekv32pushingfrontieropen,
      title={DeepSeek-V3.2: Pushing the Frontier of Open Large Language Models}, 
      author={DeepSeek-AI and Aixin Liu and Aoxue Mei and Bangcai Lin and Bing Xue and Bingxuan Wang and Bingzheng Xu and Bochao Wu and Bowei Zhang and Chaofan Lin and Chen Dong and Chengda Lu and Chenggang Zhao and others},
      year={2025},
      eprint={2512.02556},
      archivePrefix={arXiv},
      primaryClass={cs.CL},
      url={https://arxiv.org/abs/2512.02556}, 
}

@misc{comanici2025gemini25pushingfrontier,
      title={Gemini 2.5: Pushing the Frontier with Advanced Reasoning, Multimodality, Long Context, and Next Generation Agentic Capabilities}, 
      author={Gheorghe Comanici and Eric Bieber and Mike Schaekermann and Ice Pasupat and Noveen Sachdeva and Inderjit Dhillon and Marcel Blistein and Ori Ram and Dan Zhang and Evan Rosen and Luke Marris and others},
      year={2025},
      eprint={2507.06261},
      archivePrefix={arXiv},
      primaryClass={cs.CL},
      url={https://arxiv.org/abs/2507.06261}, 
}

@inproceedings{
deng2026on,
title={On the Effect of Negative Gradient in Group Relative Deep Reinforcement Optimization},
author={Wenlong Deng and Yi Ren and Muchen Li and Danica J. Sutherland and Xiaoxiao Li and Christos Thrampoulidis},
booktitle={The Thirty-ninth Annual Conference on Neural Information Processing Systems},
year={2026},
url={https://openreview.net/forum?id=2K9QsDaqkM}
}

@misc{deng2026grouprelativepolicyoptimization,
      title={On Group Relative Policy Optimization Collapse in Agent Search: The Lazy Likelihood-Displacement}, 
      author={Wenlong Deng and Yushu Li and Boying Gong and Yi Ren and Christos Thrampoulidis and Xiaoxiao Li},
      year={2026},
      eprint={2512.04220},
      archivePrefix={arXiv},
      primaryClass={cs.CL},
      url={https://arxiv.org/abs/2512.04220}, 
}

@inproceedings{li-etal-2024-dawn,
    title = "The Dawn After the Dark: An Empirical Study on Factuality Hallucination in Large Language Models",
    author = "Li, Junyi  and
      Chen, Jie  and
      Ren, Ruiyang  and
      Cheng, Xiaoxue  and
      Zhao, Xin  and
      Nie, Jian-Yun  and
      Wen, Ji-Rong",
    editor = "Ku, Lun-Wei  and
      Martins, Andre  and
      Srikumar, Vivek",
    booktitle = "Proceedings of the 62nd Annual Meeting of the Association for Computational Linguistics (Volume 1: Long Papers)",
    month = aug,
    year = "2024",
    address = "Bangkok, Thailand",
    publisher = "Association for Computational Linguistics",
    url = "https://aclanthology.org/2024.acl-long.586/",
    doi = "10.18653/v1/2024.acl-long.586",
    pages = "10879--10899"
}

@inproceedings{
he2026how,
title={How Far Can Unsupervised {RLVR} Scale {LLM} Training?},
author={Bingxiang He and Yuxin Zuo and Zeyuan Liu and Shangziqi Zhao and Zixuan Fu and Junlin Yang and Cheng Qian and Kaiyan Zhang and Yuchen Fan and Ganqu Cui and Xiusi Chen and Youbang Sun and Xingtai Lv and Xuekai Zhu and Li Sheng and Ran Li and Huan-ang Gao and Yuchen Zhang and Lifan Yuan and Bowen Zhou and Zhiyuan Liu and Ning Ding},
booktitle={The Fourteenth International Conference on Learning Representations},
year={2026},
url={https://openreview.net/forum?id=VesLZukY5E}
}

@misc{zhou2026lrasadvancedlegalreasoning,
      title={LRAS: Advanced Legal Reasoning with Agentic Search}, 
      author={Yujin Zhou and Chuxue Cao and Jinluan Yang and Lijun Wu and Conghui He and Sirui Han and Yike Guo},
      year={2026},
      eprint={2601.07296},
      archivePrefix={arXiv},
      primaryClass={cs.AI},
      url={https://arxiv.org/abs/2601.07296}, 
}

@article{bai2026glance,
  title={Glance-or-Gaze: Incentivizing LMMs to Adaptively Focus Search via Reinforcement Learning},
  author={Bai, Hongbo and Zhou, Yujin and Wu, Yile and Chan, Chi-Min and Wen, Pengcheng and Pan, Kunhao and Han, Sirui and Guo, Yike},
  journal={arXiv preprint arXiv:2601.13942},
  year={2026}
}

@inproceedings{shiintrinsic,
  title={Intrinsic Entropy of Context Length Scaling in LLMs},
  author={Shi, Jingzhe and Ma, Qinwei and Liu, Hongyi and Zhao, Hang and Hwang, Jenq-Neng and Li, Lei},
  booktitle={The Fourteenth International Conference on Learning Representations},
  year={2026}
}

@inproceedings{yao2025countllm,
title={CountLLM: Towards Generalizable Repetitive Action Counting via Large Language Model},
author={Yao, Ziyu and Cheng, Xuxin and Huang, Zhiqi and Li, Lei},
booktitle={Proceedings of the IEEE/CVF Conference on Computer Vision and Pattern Recognition (CVPR)},
year={2025}
}

@inproceedings{li2026multiple,
  title={Multiple Human Motion Understanding},
  author={Li, Lei and Jia, Sen and Hwang, Jenq-Neng},
  booktitle={Proceedings of the AAAI Conference on Artificial Intelligence},
  volume={40},
  number={8},
  pages={6297--6305},
  year={2026}
}

@InProceedings{Yan_2026_WACV,
    author    = {Yan, Ziyang and Shao, Yihua and Liao, Minwen and Chen, Siyu and Wang, Nan and Lin, Muyuan and Hwang, Jenq-Neng and Zhao, Hao and Remondino, Fabio and Li, Lei},
    title     = {3DSceneEditor: Controllable 3D Scene Editing with Gaussian Splatting},
    booktitle = {Proceedings of the IEEE/CVF Winter Conference on Applications of Computer Vision (WACV)},
    month     = {March},
    year      = {2026},
    pages     = {1852-1863}
    }

@article{zhang2026psgs,
  title={PSGS: Text-driven Panorama Sliding Scene Generation via Gaussian Splatting},
  author={Zhang, Xin and Chen, Shen and Zhou, Jiale and Li, Lei},
  journal={arXiv preprint arXiv:2602.00463},
  year={2026}
}
